\documentclass[10pt,twocolumn,letterpaper]{article}

\usepackage{xcolor,colortbl,colortab}

\usepackage{iccv}
\usepackage{times}
\usepackage{epsfig}
\usepackage{graphicx}
\usepackage{amsmath}
\usepackage{amssymb}

\usepackage{caption}
\usepackage{subcaption}
\usepackage{bbm}
\usepackage{booktabs}
\usepackage{wrapfig}

\usepackage[pagebackref=true,breaklinks=true,letterpaper=true,colorlinks,bookmarks=false]{hyperref}

\iccvfinalcopy % *** Uncomment this line for the final submission

\newcommand{\PP}{\mathbb{P}}

\usepackage{amsmath,amsfonts,bm}

\def\1{\bm{1}}

\DeclareMathAlphabet{\mathsfit}{\encodingdefault}{\sfdefault}{m}{sl}
\SetMathAlphabet{\mathsfit}{bold}{\encodingdefault}{\sfdefault}{bx}{n}

\newcommand{\sv}{{\boldsymbol s}}

\newcommand{\xv}{{\boldsymbol x}}

\def\iccvPaperID{3008} % *** Enter the ICCV Paper ID here

\ificcvfinal\fi

\begin{document}

%%%%%%%%% TITLE
\title{Proactive Pseudo-Intervention:  \\Contrastive Learning For Interpretable Vision Models}

% \author{Dong Wang, Yuewei Yang, Chenyang Tao\\
% Duke University\\
% {\tt\small dong.wang363, yuewei.yang, chenyang.tao@duke.edu}
% % For a paper whose authors are all at the same institution,
% % omit the following lines up until the closing ``}''.
% % Additional authors and addresses can be added with ``\and'',
% % just like the second author.
% % To save space, use either the email address or home page, not both
% \and
% Zhe Gan\\
% Microsoft\\
% First line of institution2 address\\
% {\tt\small secondauthor@i2.org}
% }

\author{Dong Wang$^1$, Yuewei Yang$^1$, Chenyang Tao$^1$, Zhe Gan$^2$, Liqun Chen$^1$, \\ Fanjie Kong$^1$, Ricardo Henao$^1$, Lawrence Carin$^1$\\
$^1$ Duke University, \quad $^2$ Microsoft Corporation\\
{\tt\small \{dong.wang363, yuewei.yang, chenyang.tao, liqun.chen, ricardo.henao, lcarin\}@duke.edu}, \\
{\tt\small  zhe.gan@microsoft.com}
}

\maketitle
% Remove page # from the first page of camera-ready.
\ificcvfinal\thispagestyle{empty}\fi

%%%%%%%%% ABSTRACT
\begin{abstract}
%   Deep neural networks have shown significant promise in comprehending complex visual signals, delivering performance on par or even superior to that of human experts. However, these models often lack a mechanism for interpreting their predictions, and in some cases, particularly when the sample size is small, existing deep learning solutions tend to capture spurious correlations that compromise model generalizability on unseen inputs. In this work, we propose a contrastive causal representation learning strategy that leverages proactive interventions to identify causally-relevant image features, called {\it Proactive Pseudo-Intervention} (PPI). This approach is complemented with a causal salience mapping module that identifies important pixels in the raw input image, which greatly facilitates the interpretability of predictions. To validate its utility, our model is benchmarked extensively on both standard natural images and challenging medical image datasets. We show this new contrastive causal representation learning model consistently improves model performance relative to competing solutions, particularly for out-of-domain predictions or when dealing with data integration from heterogeneous sources. Further, our causal saliency maps are more succinct and meaningful relative to their non-causal counterparts.
Deep neural networks excel at comprehending complex visual signals, delivering on par or even superior performance to that of human experts. However, ad-hoc visual explanations of model decisions often reveal an alarming level of reliance on exploiting non-causal visual cues that strongly correlate with the target label in training data. As such, deep neural nets suffer compromised generalization to novel inputs collected from different sources, and the reverse engineering of their decision rules offers limited interpretability. To overcome these limitations, we present a novel contrastive learning strategy called {\it Proactive Pseudo-Intervention} (PPI) that leverages proactive interventions to guard against image features with no causal relevance. We also devise a novel causally informed salience mapping module to identify key image pixels to intervene, and show it greatly facilitates model interpretability. To demonstrate the utility of our proposals, we benchmark on both standard natural images and challenging medical image datasets. PPI-enhanced models consistently deliver superior performance relative to competing solutions, especially on out-of-domain predictions and data integration from heterogeneous sources. Further, our causally trained saliency maps are more succinct and meaningful relative to their non-causal counterparts.
\end{abstract}

%%%%%%%%% BODY TEXT
\section{Introduction}

\begin{figure}[t!]
\centering
\includegraphics[width=0.45\textwidth]{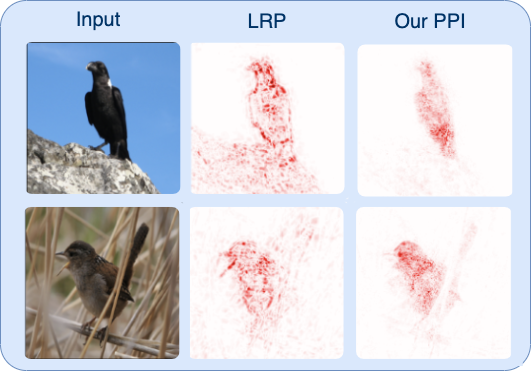}
\caption{
Interpretation for the bird-classification models using saliency maps generated by LRP ({\it layer-wise relevance propagation}) and our model PPI.
% Model interpretation for the bird-classification models using saliency maps generated by the state-of-the-art {\it layer-wise relevance propagation} (LRP) and compared to our proposed PPI.
LRP shows that naively trained deep model makes decisions based on the background cues (habitat, {\it e.g.}, rocks, bulrushes) that are spuriously correlated with the bird species, while our causally informed PPI mostly focuses on the bird anatomy, that generalizes beyond the natural habitat.}
\label{fig:examples_intro}
\vspace{-1em}
\end{figure}

Deep neural networks hold great promise in applications requiring the analysis and comprehension of complex imagery.
Recent advances in hardware, network architectures, and model optimization, along with the increasing availability of large-scale annotated datasets~\cite{krizhevsky2009learning,deng2012mnist,deng2009imagenet}, have enabled these models to match and sometimes outperform human experts on a number of tasks, including natural image classification~\cite{krizhevsky2017imagenet}, objection recognition~\cite{girshick2014rich}, disease diagnosis~\cite{sajda2006machine}, and autonomous driving~\cite{chen2015deepdriving}, among others.

While deep learning solutions have been positively recognized for their ability to learn {\em black-box} models in a purely data driven manner, their very nature makes them less credible for their inability to communicate the reasoning for making predictions in a way that is comprehensible to humans~\cite{hooker2019benchmark, rebuffi2020there}. This denies consequential applications where the reliability and trustworthiness of a prediction are of primary concern and require expert audit, {\it e.g.}, in healthcare~\cite{sajda2006machine}.
To stimulate widespread use of deep learning models, a means of interpreting predictions is necessary. However, model interpretation techniques often reveal a concerning fact, that deep learning models tend to assimilate spurious correlations that do not necessarily capture the causal relationship between the input (image) and output (label)~\cite{wang2020score}. This issue is particularly notable in small-sample-size (weak supervision) scenarios or when the sources of non-informative variation are overwhelming, thus likely to cause severe overfitting.
These can lead to catastrophic failures on deployment~\cite{fukui2019attention, wang2019sharpen}.
% Such concerns are actually very well justified, in the absence of an educated critic, unprotected deep models tend to assimilate spurious correlations rather than to capture the true underlying causal relations.
% Such negligence, albeit unintentional, can often lead to catastrophic failures on deployment, when processing inputs that are different from the training examples. 

A growing recognition of the issues associated with the lack of interpretable predictions is well documented in recent years \cite{adebayo2018sanity, hooker2019benchmark, rebuffi2020there}.
Such phenomenon has energized researchers to actively seek creative solutions.
Among these, two streams of work, namely {\it saliency mapping}~\cite{zhao2018uniqueness, simonyan2013deep, dabkowski2017real} and {\it causal representation learning} (CRL)~\cite{johansson2016learning, wang2020visual, arjovsky2019invariant}, stand out as some of the most promising directions.
Specifically, saliency mapping encompasses techniques for {\it post hoc} visualizations on the input (image) space to facilitate the interpretation of model predictions.
This is done by projecting the key features used in prediction back to the input space, resulting in the commonly known {\it saliency maps}.
% These techniques typically rely on the notion of sensitivity of the input to changes in the model output, to derive a map of image regions that are important for prediction.
Importantly, these maps do not directly contribute to model learning.
% ping for expert evaluation, and do not directly contribute to the model training procedure.
Alternatively, CRL solutions are built on the principles of establishing invariance from the data, and it entails teasing out sources of variation that are spuriously associated with the model output (labels).
% often necessitates the identification of environments to disentangle nuisance heterogeneities that contributes to spurious associations.
CRL models, while emphasizing the differences between causation and correlation, are not subject to the rigor of causal inference approaches, because their goal is not to obtain accurate causal effect estimates but rather to produce robust models with better generalization ability relative to their naively learned counterparts \cite{arjovsky2019invariant}.
% do not actually subject themselves to expert scrutiny for validation of their causal findings 

In this work, we present {\it Proactive Pseudo-Intervention} (PPI), a solution that accounts for the needs of causal representation identification and visual verification.
Our key insight is the derivation of causally-informed saliency maps, which facilitate visual verification of model predictions {\em and} enable learning that is robust to (non-causal) associations.
While true causation can only be established through experimental interventions, we leverage tools from contrastive representation learning to synthesize pseudo-interventions from observational data. Our procedure is motivated by the causal argument: perturbing the non-causal features will not change the target label. 
% the target label will be changed only if causally-relevant features are perturbed.

To motivate, in Figure \ref{fig:examples_intro} we present an example to illustrate the benefits of producing causally-informed saliency maps.
% To provide a concrete example of our causal reasoning, see Figure \ref{fig:examples_intro} for an intuitive illustration.
% A is Groove-billed ani, B is Marsh Wren.
In this scenario, the task is to classify two bird species (A and B) in the wild.
Due to the differences in their natural habitats, A-birds are mostly seen resting on rocks, while B-birds are more commonly found among bulrushes. % Marsh Wren breeds in many fresh and brackish marsh situations, usually with a large area of cattails, bulrushes, or cordgrass %foraging on the ground.
A deep model, trained naively, will tend to associate the background characteristics with the labels, knowing these strongly correlate with the bird species (labels) in the training set.
This is confirmed by the saliency maps derived from the layer-wise relevance propagation (LRP) techniques \cite{bach2015pixel}: the model also attends heavily on the background features, while the difference in bird anatomy is what causally determines the label.
If we were provided with an image of a bird in an environment foreign to the images in the training set, the model will be unable to make a reliable prediction, thus causing robustness concerns.
% (or if  {\em different} types of animals were in the same environments, they could be confused with bird A or B).
This generalization issue worsens with a smaller training sample size.
On the other hand, saliency maps from our PPI-enhanced model successfully focus on the bird anatomy, and thus will be robust to environmental changes captured in the input images.
% Alternatively, a causally-informed module of our PPI, like the proposed Weight Back Propagation (WBP), will successfully focus on the bird anatomy, and thus will be robust to environmental changes captured in the input images.

%This paper presents an easy-to-implement strategy called {\it Proactive Pseudo-Intervention} (PPI) that 
PPI addresses causally-informed reasoning, robust learning, and model interpretation in a unified framework. 
A new saliency mapping method, named {\it Weight Back Propagation} (WBP), is also proposed to generate more concentrated intervention mask for PPI training.
%PPI seamlessly combines saliency mapping and contrastive interventions to guide model learning.
The key contributions of this paper include:
\begin{itemize}
    \vspace{-5pt}
    \item An end-to-end contrastive representation learning strategy PPI that employs proactive interventions to identify causally relevant features.
    \vspace{-5pt}

    \item A fast and architecture-agnostic saliency mapping module WBP that delivers better visualization and localization performance.% accurate and faithful interpretation of predictions.  
    \vspace{-5pt}
    
    \item Experiments demonstrating significant performance boosts from integrating PPI and WBP relative to competing solutions, especially on  out-of-domain predictions, data integration with heterogeneous sources and model interpretation.
\end{itemize}

\section{Background}
% ---- New version
\noindent{\bf Visual Explanations}
Saliency mapping collectively refers to a family of techniques to understand and interpret black-box image classification models, such as deep neural networks~\cite{adebayo2018sanity, hooker2019benchmark, rebuffi2020there}.
These methods project the model understanding of the targets, {\it i.e.}, labels, and their predictions back to the input space, which allows for the visual inspection of automated reasoning and for the communication of predictive visual cues to the user or human expert, aiming to shed model insights or to build trust for deep-learning-based systems. 
% , {\it e.g.}, model diagnosis and data insights, to human experts, to establish the reliability and build trust in AI systems.

In this study, we focus on {\it post hoc} saliency mapping strategies, where saliency maps are constructed given an arbitrary prediction model, as opposed to relying on customized model architectures for interpretable predictions~\cite{fukui2019attention,wang2019sharpen}, or to train a separate module to explicitly produce model explanations \cite{fukui2019attention, goyal2019counterfactual, chang2018explaining,  fong2017interpretable, shrikumar2017learning}.
Popular solutions under this category include: activation mapping~\cite{zhou2016learning,selvaraju2017grad}, input sensitivity analysis~\cite{shrikumar2017learning}, and relevance propagation \cite{bach2015pixel}. 
Activation mapping based methods fail at visualizing fine-grained evidence, which is particularly important in explaining medical classification models~\cite{du2018towards, selvaraju2017grad, wagner2019interpretable}. Input sensitivity analysis based methods produce fine-grained saliency maps. However, these maps are generally less concentrated~\cite{dabkowski2017real,fong2017interpretable} and less interpretable. 
Relevance propagation based methods, like LRP and its variants, use complex rules to prioritize positive or large relevance, making the saliency maps visually appealing to human. However, our experiments demonstrate that LRP and its variants highlight spuriously correlated features (boarderlines and backgrounds). 
By contrast, our WBP backpropagates the weights through layers to compute the contributions of each input pixel, which is truly faithful to the model, and WBP tends to highlight the target objects themselves rather than the background. At the same time, the simplicity and efficiency makes WBP easily work with other advanced learning strategies for both model diagnosis and improvements during training. 
%As we will show in the experiments, these approaches struggle to yield causal explanations of the outcome, and are therefore unable to satisfactorily address the challenges discussed above. 

Our work is in a similar spirit to \cite{fong2017interpretable, dabkowski2017real, chang2018explaining, wagner2019interpretable}, where meaningful perturbations have been applied to the image during model training, to improve prediction and facilitate interpretation. Poineering works have relied on user supplied ``ground-truth'' explainable masks to perturb \cite{ross2017right,li2018tell, rieger2020interpretations}, however such manual annotations are costly and hence rarely available in practice. Alternatively, perturbations can be computed by solving an optimization for each image. Such strategies are costly in practice and also do not effectively block spurious features.  Very recently, exploratory effort has been made to leverage the tools from counterfactual reasoning \cite{goyal2019counterfactual} and causal analysis \cite{o2020generative} to derive visual explanations, but do not lend insights back to model training. Our work represents a fast, principled solution that overcomes the above limitations. It automatically derives explainable masks faithful to the model and data, without explicit supervision from user-generated explanations.

% -- same
\vspace{2mm}
\noindent{\bf Contrastive Learning. }
There has been growing interest in exploiting contrastive learning (CL) techniques for representations learning \cite{oord2018representation,chen2020simple,he2020momentum,khosla2020supervised,tian2019contrastive}.
Originally devised for density estimation \cite{gutmann2010noise}, CL exploits the idea of {\em learning by comparison} to capture the subtle features of data, {\it i.e.}, positive examples, by contrasting them with negative examples drawn from a carefully crafted noise distribution.
These techniques aim to avoid representation collapse, or to promote representation consistency, for downstream tasks.
Recent developments, both empirical and theoretical, have connected CL to information-theoretic foundations~\cite{tian2019contrastive,grill2020bootstrap}, thus establishing them as a suite of {\it de facto} solutions for unsupervised representation learning \cite{chen2020simple,he2020momentum}. 

The basic form of CL is essentially a binary classification task specified to discriminate positive and negative examples.
In such a scenario, the binary classifier is known as the critic function.
Maximizing the discriminative power wrt the critic and the representation sharpens the feature encoder. 
Critical to the success of CL is the choice of appropriate noise distribution, where the challenging negatives, {\it i.e.}, those negatives that are more similar to positive examples, are often considered more effective contrasts. 
In its more generalized form, CL can naturally repurpose the predictor and loss functions without introducing a new critic~\cite{tian2019contrastive}.
Notably, current CL methods are not immune to spurious associations, a point we wish to improve in this work.

\vspace{2mm}

%---- New version
\noindent{\bf Causality and Interventions.} From a causality perspective, humans learn via actively interacting with the environment. We intervene and observe changes in the outcome to infer causal dependencies. Machines instead learn from static observations that are unable to inform the structural dependencies for causal decisions. As such, perturbations to the external factors, {\it e.g.}, surroundings, lighting, viewing angles, may drastically alter machine predictions, while human recognition is less susceptible to such nuisance variations.
% The success of many powerful computer vision models often hinges on the over-exploitation of visual cues.
% Perturbations to the external factors, {\it e.g.}, surroundings, lighting, viewing angles, may drastically alter their predictions, while human recognition is less susceptible to such nuisance variations.
% % Rooting the cause
% This happens because standard machine learning models base their decision on correlations, as opposed to performing causal reasoning as humans do.
% observational data are in general affected by (unknown) confounding factors, and consequently are unable to inform the structural dependencies for causal decisions.
Formally, such difference is best explained with the $do$-notation~\cite{pearl2009causality}: $\PP(Y|do(x)) = \sum_z \PP(Y|X=x, z) \PP(z)$, where we identify $x$ as the features, {\it e.g.}, an object in the image, and $z$ as the confounders, {\it e.g.}, background in the example above.
% In the causality literature, $x$ is commonly referred to as the treatment~\cite{arjovsky2019invariant, ghassami2017learning}.
Note that $\PP(Y|do(x))$ is fundamentally different from the conditional likelihood $\PP(Y|X=x) = \sum_z \PP(Y|X=x,z) \PP(z|X=x)$, which machine uses for associative reasoning.
% Employing the {\it do}-interventions explicitly alter the causal graph, which breaks the spurious correlations and leaves true causal relations. 
%$\PP(Y|do(x)) \neq \PP(Y|x) = \sum_z \PP(Y|X=x,z) \PP(z|X=x)$. 
%\begin{eqnarray}
%\PP(Y|x) = \sum_z \PP(Y|X=x,z) \PP(z|X=x), \\
%\text{{\bf Causal:} } \PP(Y|do(x)) = \sum_z \PP(Y|X=x, z) \PP(z), 
%\end{eqnarray}

% Acquiring causal knowledge requires experimental {\em interventions} to intentionally block non-causal associations.
% This action is typically implemented by explicitly intervening the experiments by fixing $x$, and then observing how it affects the outcome $y$.
Unfortunately, carrying out real interventional studies, {\it i.e.}, randomized control trials, to intentionally block non-causal associations, is oftentimes not a feasible option for practical considerations, {\it e.g.}, due to cost and ethics.
This work instead advocates the application of synthetic interventions to uncover the underlying causal features from observational data.
Specifically, we proactively edit $x$ and its corresponding label $y$ in a data-driven fashion to encourage the model to learn potential causal associations. Our proposal is in line with the growing appreciation for the significance of establishing causality in machine learning models \cite{scholkopf2019causality}. Via promoting invariance \cite{arjovsky2019invariant}, such causally inspired solutions demonstrate superior robustness to superficial features that do not generalize \cite{wang2019learning}. In particular, \cite{suter2019robustly, zhang2020causal} showed the importance and effectiveness of accounting for interventional perspectives. Our work brings these causal views to construct a simple solution that explicitly optimizes visual interpretation and model robustness. 

\begin{figure*}[h!]
    \centering
    \includegraphics[width=0.98\textwidth]{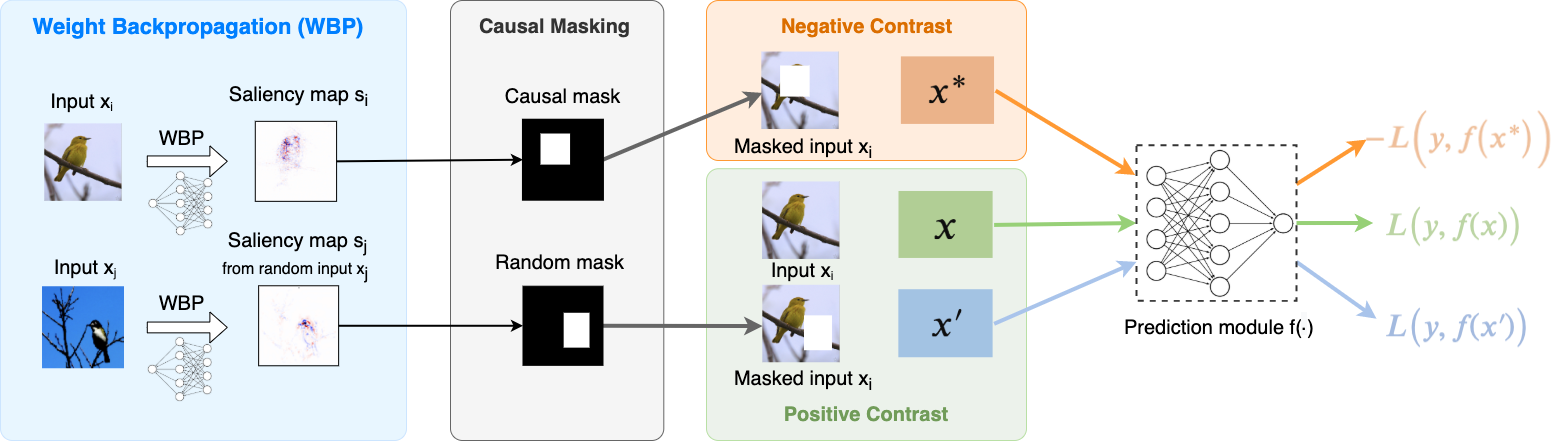}%,trim=0 0 0 20mm, clip]{./Figures/learning_framework_1116_l.png} %{./Figures/learning_1121.png}
    \caption{Illustration of the proposed PPI learning strategy. 
    Input images are intervened by removing the saliency map based masks, which alters the input label ({\it e.g.}, negative control). For positive contrast, we use the original input as well as an input masked with a random slaiency map. We use WBP for the generation of saliency maps. 
    % Saliency maps are obtained via WBP (Left), contrastive interventions are created from masked inputs (Middle), and predictions are obtained from a classification model (Right).
    }
    \label{fig:learning_strategy}
    \vspace{-1em}
\end{figure*}

\vspace{-3pt}
\section{Proactive Pseudo-Intervention}
Below we describe the construction of {\it Proactive Pseudo-Intervention} (PPI), a causally-informed contrastive learning scheme that seeks to simultaneously improve the accuracy, robustness, generalization and interpretability of deep-learning-based computer vision models. 

The PPI learning strategy, schematically summarized in Figure ~\ref{fig:learning_strategy}, consists of three main components: ($i$) a saliency mapping module that highlights causally relevant features; ($ii$) an intervention module that synthesizes contrastive samples; and ($iii$) the prediction module, which is standard in recent vision models, {\it e.g.}, VGG~\cite{simonyan2014very}, ResNet~\cite{he2016deep}, and Inception Net~\cite{szegedy2016rethinking}. 
Motivated by the discussions from our introduction, PPI establishes a feedback loop between the saliency map module and the prediction module, which is interfaced by the synthesized contrastive examples in the intervention module.
Under this configuration, the prediction module is encouraged to modify its predictions only when provided with causally-relevant synthetic interventions.
% The key consideration being that only causally relevant synthetic interventions can alter the predicted outcomes, thus conforming to this rule sharpens both the salience maps and prediction.
Note that components ($i$) and ($ii$) do not involve any additional parameters or neural network modules, which makes our strategy readily applicable to the training of virtually any computer vision task without major customization.
Details of these building blocks are given below.

\vspace{-2pt}
\subsection{Synthetic causal interventions for contrasts}
Key to our formulation is the design of a synthetic intervention strategy that generates contrastive examples to reinforce causal relevance during model training.
Given a causal saliency map $\sv_m(\xv)$ for an input $\xv$ wrt label $y=m$, where $m=1,\ldots,M$, and $M$ is the number of classes, the synthetic intervention consists of removing (replacing with zero) the causal information from $\xv$ contained in $\sv_m(\xv)$, and then using it as the contrastive learning signal.
% we remove (zero-out) the causal information from the input as our synthetic intervention, and then use it as the contrastive learning signal.

\begin{figure*}[t!]
\centering
\includegraphics[width=0.97\textwidth]{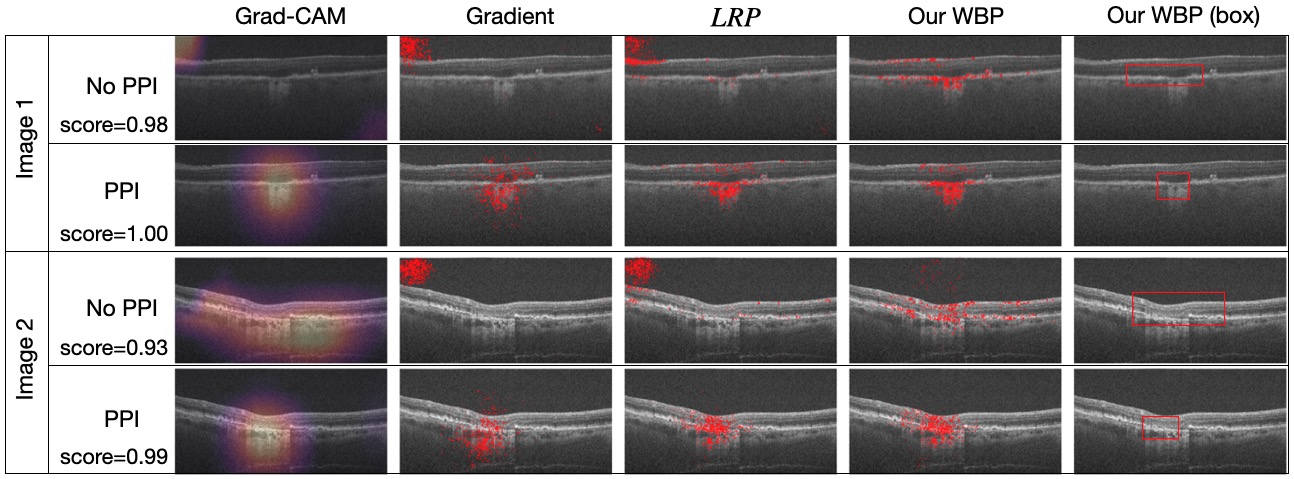}%{./Figures/exp_ga_1_img.png} %{./Figures/exp_ga_1.png} 
\caption{Saliency maps on GA dataset based on models trained with PPI and without PPI. Maps of models trained with PPI are more clinically relevant by focusing on retinal layers likely to contain abnormalities or lesions, and more concentrated. }
\label{fig:exp_ga_1}
\vspace{-1em}
\end{figure*}

For now, let us assume the causal salience map $\sv_m(\xv)$ is known; the procedure to obtain the saliency map will be addressed in the next section.
For notational clarity, we use subscript $i$ to denote entities associated with the $i$-th training sample, and omit the dependency on learnable parameters.
To remove causal information from $\xv_i$ and obtain a negative contrast $\xv_i^{*}$, we apply the following {\it soft-masking} 
\begin{equation}\label{eq:mask}
    \xv_i^{*} = \xv_i - T(\sv_m(\xv_i)) \odot \xv_i,
\end{equation}
where $T(\cdot)$ is a differentiable masking function and $\odot$ denotes element-wise (Hadamard) multiplication.
Specifically, we use the thresholded sigmoid for masking:
\begin{equation}\label{eq:sig}
    T(\sv_m(\xv_i)) = \frac{1}{1+\exp(-\omega (\sv_m(\xv_i) - \mathbf{\sigma}))}, 
\end{equation}
where $\sigma$ and $\omega>0$ are the threshold and scaling parameters, respectively.
We set the scaling $\omega$ so that $T(\sv)$ will result in a sharp transition from $0$ to $1$ near $\sigma$.
Using \eqref{eq:mask} we define the contrastive loss as
\begin{equation}\label{loss_fn1}
    L_{con}(\theta) = \sum_i \ell (\xv_i^{*}, \neg y; f_{\theta}),
\end{equation}
where $f_{\theta}$ is the prediction module, $\ell(\xv, y; f_{\theta})$ is the loss function we wish to optimize, {\it e.g.} cross entropy, and $\neg$ is used to denote that the original class label has been flipped.
In the binary case, $\neg y= 1-y$, and in the multi-class case it can be interpreted accordingly, {\it e.g.}, using a one {\it vs}. others cross entropy loss. 
In practice, we set $\ell(\xv, y; f_{\theta}) = - \ell(\xv, y; f_{\theta})$.
We will show in the experiments 
that this simple and intuitive causal masking strategy works well in practice (see Tables  \ref{tab:exp_ga} and \ref{tab:exp_lidc}, and Figure~\ref{fig:exp_ga_1}). 
Alternatively, we also consider a {\it hard-masking} approach in which a minimal bounding box covering the thresholded saliency map is removed.
See the Appendix for details.

Note that we are making the implicit assumption that the saliency map is uniquely determined by the prediction module $f_{\theta}$.
While optimizing \eqref{loss_fn1} explicitly attempts to improve the fit of the prediction module $f_{\theta}$, it also implicitly informs the causal saliency mapping.
This is sensible because if a prediction is made using non-causal features, which implies the associated saliency map $\sv_m(\xv)$ is also non-causal, then we should expect that after applying $\sv_m(\xv)$ to $\xv$ using \eqref{eq:mask}, we can still expect to make the correct prediction, {\it i.e.}, the true label, for both positive (the original) and negative (the intervened) samples.

% \begin{figure*}[h!]
%     \centering
%     \includegraphics[width=0.95\textwidth]{./Figures/wbp_allin1_tight.png}
%     \caption{Illustration of the Weight Backpropagation (WBP) through a fully connected layer with ReLU activation layer.}
%     \label{fig:wb_pool}
%     \vspace{-1em}
%  \end{figure*}

\vspace{1mm}
\noindent {\bf Saliency map regularization.}
Note that naively optimizing \eqref{loss_fn1} can lead to degenerate solutions for which any saliency map that satisfies the causal sufficiency, {\it i.e.}, encompassing all causal features, is a valid causal saliency map.
For example, a trivial solution where the saliency map covers the entire image may be considered causal.
% of all $1$s in the saliency map may be considered causal.
To protect against such degeneracy, we propose to regularize the $L_1$-norm of the saliency map to encourage succinct (sparse) representations, {\it i.e.}, $L_{reg} = \| \sv_m \|_1$, for $m=1,\ldots,M$.

\vspace{1mm}
\noindent {\bf Adversarial positive contrasts.}
Another concern with solely optimizing \eqref{loss_fn1} is that models can easily overfit to the intervention, {\it i.e.}, instead of learning to capture causal relevance, the model learns to predict interventional operations.
For example, the model can learn to change its prediction when it detects that the input has been intervened, regardless of whether the image is missing causal features.
So motivated, we introduce adversarial positive contrasts:
\begin{equation}\label{eq:random}
    \xv'_i = \xv_i - T(\sv_m(\xv_j)) \odot \xv_i, \quad i \neq j, 
\end{equation}
where we intervene with a {\em false} saliency map, {\it i.e.}, $\sv_m(\xv_j)$ is the saliency map from a different input $\xv_j$, while still encouraging the model to make the correct prediction via
% that is to say, apply a false mask for intervention, and the model is still incentivized to make the right prediction 
%
\begin{equation}\label{loss_fn2}
    L_{ad}(\theta) = \sum_i \ell (\xv_i', y ; f_{\theta})\,,
    \vspace{-3mm}
\end{equation}
where $\xv'_i$ is the adversarial positive contrast. 
The complete loss for the proposed model, $L = L_{cls} + L_{con} + L_{reg} + L_{ad}$, consists of the contrastive loss in \eqref{loss_fn1}, the regularization loss, $L_{reg}$, and the adversarial loss in \eqref{loss_fn2}.

\subsection{Saliency Weight Backpropagation}
In order to generate saliency maps that inform decision-driving features in the (raw) pixel space, we describe Weight Back Propagation (WBP), a novel computationally efficient scheme for saliency mapping applicable to arbitrary neural architectures. WBP evaluates individual contributions from each pixel to the final class-specific prediction, and we empirically find the results to be more causally-relevant relative to competing solutions based on human judgement. 

\begin{table}[t!]
    \caption{WBP update rules for common transformations.}
    \centering
    \begin{tabular}{ll}
    \toprule 
    Transformation & $G(\cdot)$ \\ \midrule
    \textit{Activation Layer} & $\tilde{W}^l=h\circ{\tilde{W}^{l+1}}$\\ %\hline
    \textit{FC Layer} & $\tilde{W}^{l}=\tilde{W}^{l+1}W^{l}$\\ %\hline
    \textit{Convolutional Layer} & $\tilde{W}^{l}=\tilde{W}^{l+1}\otimes{[W^{l}]^{T_{0,1}}_{flip_{2,3}}}$\\ %\hline
    \textit{BN Layer} & $\tilde{W}^l=\frac{\tilde{W}^{l+1}}{\sigma}\gamma$ \\ %\hline
    \textit{Pooling Layer}  & Relocate/Distribute $\tilde{W}^{l+1}$\\ \bottomrule
    \end{tabular}
    \label{tbl:wbp_layers}
 \end{table}

To simplify our presentation, we first consider a vector input and a linear mapping. 
Let $\xv^l$ be the internal representation of the data at the $l$-th layer, with $l=0$ being the input layer, {\it i.e.}, $\xv^0 = \xv$, and $l=L$ being the penultimate {\it logit} layer prior to the softmax transformation, {\it i.e.}, $\PP(y|\xv) = {\rm softmax}(\xv^L)$.
To assign the relative importance to each hidden unit in the $l$-th layer, we notationally collapse all transformations after $l$ into an operator denoted by $\tilde{W}^l$, which we call the {\it saliency matrix}, satisfying,
\begin{equation}
    \xv^L = \tilde{W}^l \xv^l, \quad \forall l\in [0, \ldots, L], 
\end{equation}
where $\xv^L$ is an $M$-dimensional vector corresponding to the $M$ distinct classes in $y$.
Though presented in a matrix form in a slight abuse of notation, {\it i.e.}, the instantiation of the operator $\tilde{W}^l$ effectively depends on the input $\xv$, thus all nonlinearities have been effectively absorbed into it.
% It is reasonable to stipulate that,
We posit that for an object associated with a given label $y=m$, its causal features are subsumed in the interactions between the $m$-th row of $\tilde{W}^0$ and input $\xv$, {\it i.e.},
\begin{equation}
    [\sv_m(\xv)]_k = [\tilde{W}^0]_{mk} [\xv]_k, 
\end{equation}
where $[\sv_m(\xv)]_k$ denotes the $k$-th element of the saliency map $\sv_m(\xv)$ and $[\tilde{W}^0]_{mk}$ is a single element of $\tilde{W}^0$.
A key observation for computation of $\tilde{W}^l$ is that it can be done recursively.
Specifically, let $g_l(\xv^l)$ be the transformation at the $l$-th layer, {\it e.g.}, an affine transformation, convolution, activation, normalization, {\it etc}., then it holds that
\begin{equation}
    \tilde{W}^{l+1} \xv^{l+1} = \tilde{W}^{l+1} g_l(\xv^l) = \tilde{W}^{l} \xv^l. 
\end{equation}
This allows for recursive computation of $\tilde{W}^{l}$ via
\begin{equation}
    \tilde{W}^{l} = G(\tilde{W}^{l+1}, g_l), \quad \tilde{W}^{L} = 1,
\end{equation}
%
% \rhg{where $G(\cdot)$ is?} \dw{$G(\cdot)$ is the update rule listed in the table. I have updated the following sentence.}
where $G(\cdot)$ is the update rule.
We list the update rules for common transformations in deep networks in Table~\ref{tbl:wbp_layers}, with corresponding derivations detailed below.

% \begin{figure*}[thpb!]
% \centering
% \includegraphics[width=0.95\textwidth]{./Figures/exp_cub_red_x.jpg}%{./Figures/exp_cub_red.png} %{latex/Figures/exp_cub.jpg}
% \caption{Comparisons of saliency maps on CUB dataset.}
% \label{fig:exp_cub}
% \vspace{-1em}
% \end{figure*}

\vspace{2mm}
\noindent{\bf Fully-connected (FC) layer.} 
The FC transformation is the most basic operation in deep neural networks.
Below we omit the bias term as it does not directly interact with the input.
Assuming $g_l(\xv^l) = W^l \xv^l$, it is readily seen that
\begin{equation}
    \tilde{W}^{l+1} \xv^{l+1} = \tilde{W}^{l+1} g_l(\xv^l) = (\tilde{W}^{l+1} W^l) \xv^l, 
\end{equation}
so $\tilde{W}^{l} = \tilde{W}^{l+1} W^l$.
Graphical illustration with standard affine mapping and ReLU activation can be found in the appendix.
% See Figure~\ref{fig:wb_pool} for a graphical illustration with standard affine mapping and ReLU activation.

\begin{figure*}[t!]
\centering
\includegraphics[width=0.95\textwidth]{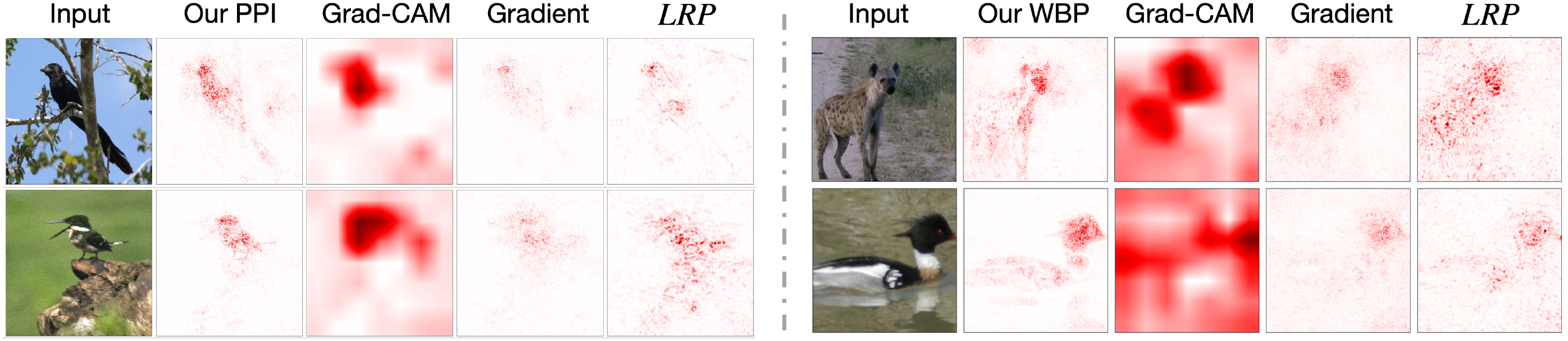}%{./Figures/exp_cub_red.png} %{latex/Figures/exp_cub.jpg}
\caption{Visualization of the inferred saliency maps. Left: CUB dataset. Right: ImageNet dataset. }
\label{fig:exp_cub}
\vspace{-1em}
\end{figure*}

\vspace{2mm}
\noindent{\bf Nonlinear activation layer.} 
Considering that an activation layer simply {\em rescales} the saliency weight matrices, {\it i.e.}, $\xv^{l+1} = g_l(\xv^l) = h^l \circ \xv^l$, where $\circ$ is the composition operator, we obtain $\tilde{W}^{l} = h \circ \tilde{W}^{l+1}$.
Using the ReLU activation as a concrete example, we have $h(\xv^l) = \mathbbm{1}\{\xv^l\geq 0\}$.

\vspace{2mm}
\noindent{\bf Convolutional layer.}
% Marking the era of deep vision models is the wide employment of convolutions for image inputs.
The convolution is a generalized form of linear mapping.
In practice, convolutions can be expressed as tensor products of the form $\tilde{W}^{l}=\tilde{W}^{l+1}\otimes{[W^{l}]^{T_{0,1}}_{flip_{2,3}}}$, where $W^l\in{\mathbb{R}^{D_2\times{D_1}\times{(2S+1)}\times{(2S+1)}}}$ is the convolution kernel, $T_{0,1}$ is the transpose in dimensions 0 and 1 and $flip_{2,3}$ is an exchange in dimensions 2 and 3.
See the Appendix for details.

\vspace{2mm}
\noindent{\bf Pooling and normalization layer.} Summarization and standardization are two other essential operations for the success of deep neural networks, achieved by pooling and batch normalization (BN) techniques, respectively.
They too can be considered as special instantiations of linear operations.
Here we summarize the two most popular operations in Table \ref{tbl:wbp_layers}. 

% \begin{figure}[thpb!]
% \centering
% \includegraphics[width=0.45\textwidth]{./Figures/exp_cub_1.jpg}%{./Figures/exp_cub_red.png} %{latex/Figures/exp_cub.jpg}
% \caption{Comparisons of saliency maps on CUB dataset.}
% \label{fig:exp_cub}
% \vspace{-1em}
% \end{figure}

% \begin{figure}[thpb!]
% \centering
% \includegraphics[width=0.45\textwidth]{./Figures/exp_imagenet_1.jpg}%{./Figures/exp_cub_red.png} %{latex/Figures/exp_cub.jpg}
% \caption{Comparisons of saliency maps on ImageNet dataset.}
% \label{fig:exp_imagenet}
% \vspace{-1em}
% \end{figure}
\section{Experiments}
To validate the utility of our approach, we consider both natural  and medical image datasets, and compare it to existing state-of-the-art solutions.
All the experiments are implemented in PyTorch.
The source code will be available at \url{https://github.com/author_name/PPI}.
Due to space limitation, details of the experimental setup and additional analyses are deferred to the Appendix.

\vspace{2mm}
\noindent{\bf Datasets.} We present our findings on five representative datasets: ($i$) \texttt{CIFAR-10}~\cite{krizhevsky2009learning}; ($ii$) \texttt{ImageNet (ILSVRC2012)}~\cite{ILSVRC15}; ($iii$) \texttt{CUB}~\cite{WahCUB_200_2011}, a natural image dataset with over $12k$ photos for classification of $200$ bird species in the wild, heavily confounded by the background characteristics; ($iv$) \texttt{GA}~\cite{leuschen2013spectral}, a new medical image dataset for the prediction of {\it geographic atrophy} (GA) using 3D {\it optical coherence tomography} (OCT) image volumes, characterized by small sample size ($275$ subjects) and highly heterogeneous (collected from $4$ different facilities); and ($v$) \texttt{LIDC-IDRI}~\cite{langlotz2019roadmap}, a public medical dataset of $1,085$ lung lesion CT images annotated by $4$ radiologists.
Detailed specifications are described in the Appendix. 

\vspace{2mm}
\noindent{\bf Baselines.} The following set of popular saliency mapping schemes are considered as comparators for the proposed approach: ($i$) Gradient: standard gradient-based salience mapping; ($ii$) GradCAM \cite{selvaraju2017grad}: gradient-weighted class activation mapping; ($iii$) LRP~\cite{bach2015pixel}: layer-wise relevance propagation and its variants. 

\vspace{2mm}
\noindent{\bf Hyperparameters.} 
The final loss of the proposed model is a weighted summation of four losses: $L = L_{cls} + w_1 L_{con} + w_2 L_{reg} + w_3 L_{ad}$.
The weights are simply balanced to match the magnitude of $L_{cls}$, \emph{i.e.}, $w_{3}=1$,  $w_{2}=0.1$ and $w_{1}=1$ (CUB), $=1$ (GA), and $=10$ (LIDC).
%To assess sensitivity, we re-scaled $w_{ad}$, $w_{reg}$ and $w_{con}$ by a factor in the set $\{0.5,1,3\}$ while keeping $w_{cls}=1$, from which we see no obvious changes in the results. 
See Appendix Sec B for details about the masking parameters $\sigma$ and $\omega$.

% The experiment part is organized as the followings: we report see consistent performance gains in classification via incorporating  PPI  training in Section~\ref{sec:classification_gain}; we discuss the performance gains in the perspective of model interpretability in Section~\ref{sec:interpretation_gain}; we also want to highlight the significant improvements on small sample size and heterogeneity in medical image datasets in Section~\ref{sec:exp_ga} and \ref{sec:exp_lidc}. 

\subsection{Natural Image Datasets}
\noindent \textbf{Classification Gains} %\label{sec:classification_gain}
In this experiment, we investigate how the different pairings of PPI and saliency mapping schemes ({\it i.e.}, GradCAM, LRP, WBP) affect performance.
In Table \ref{tab:exp_ga}, the first row represents VGG11 model trained with only classification loss, and the following rows represent VGG11 trained with PPI with different saliency mapping schemes. We see consistent performance gains in accuracy via incorporating PPI training on both CUB and CIFAR-10 datasets. The gains are mostly significant when using our WBP for saliency mapping (improving the accuracy from $0.662$ to $0.696$ on CUB, and from $0.881$ to $0.901$ on CIFAR-10.

\begin{table}[t!]
    \caption{Performance improvements achieved by training with PPI on CUB,  CIFAR-10, and GA dataset. We report means and standard deviations (SDs) from 5-fold cross-validation for GA prediction.}
    \centering
    \begin{tabular}{llll} %p{40pt}p{50pt}p{170pt}p{130pt}
        \toprule
        \textbf{Models} & {\textbf{CUB}} & \textbf{Cifar-10}  & \textbf{GA }    \\ 
        \textbf{{}} & {\textbf{(Acc)}} & \textbf{(Acc)}  & \textbf{(AUC)}    \\ \midrule
        % Random Forest~\cite{x} & 0.787 & 0.021\\ [5pt]%\hline
        Classification & $0.662$ & $0.881$ & $0.877 \pm 0.040$\\  
        +$\text{PPI}_{Gradient}$ & ${0.673}$ & $0.885$ & $0.890 \pm 0.035$\\ 
        +$\text{PPI}_{LRP}$ & ${0.680}$ & $0.891$ & $0.895 \pm 0.037$\\ 
        +$\text{PPI}_{GradCAM}$ & ${0.683}$ & $0.895$ & $0.908 \pm 0.036$ \\ 
        +$\text{PPI}_{WBP}$ & \textbf{0.696} & \textbf{0.901} & $0.925 \pm 0.023$ \\ 
        % +$\text{PPI}_{WBP (box)}$ & {-} & - & \textbf{0.937 $\pm$ 0.015}\\ 
        \bottomrule
    \end{tabular}
    \label{tab:exp_ga}
    \vspace{-4mm}
\end{table}
% \vspace{2mm}

\noindent\textbf{Model Interpretability}
% \paragraph{Model Interpretability}
%\label{sec:interpretation_gain}
In this task, we want to qualitatively and quantitatively compare the causal relevance of saliency maps generated by our proposed model and its competitors.
In Figure \ref{fig:exp_cub}, we show the saliency maps produced by different approaches for a VGG11 model trained on CUB. % (with $0.662$ testing accuracy).
Visually, gradient-based solutions (Grad and GradCAM) tend to yield overly dispersed maps, indicating a lack of specificity.
LRP gives more appealing saliency maps. However, these maps also heavily attend to the spurious background cues that presumably help with predictions. When trained with PPI, the saliency maps attend to birds body, and with WBP, the saliency maps focus on the causal related pixels.
%In contrast, WBP faithfully focuses the attention to the birds themselves.

\begin{figure}[!tbp]
\centering
\begin{subfigure}[b]{0.23\textwidth}
         \centering
         \includegraphics[width=\textwidth]{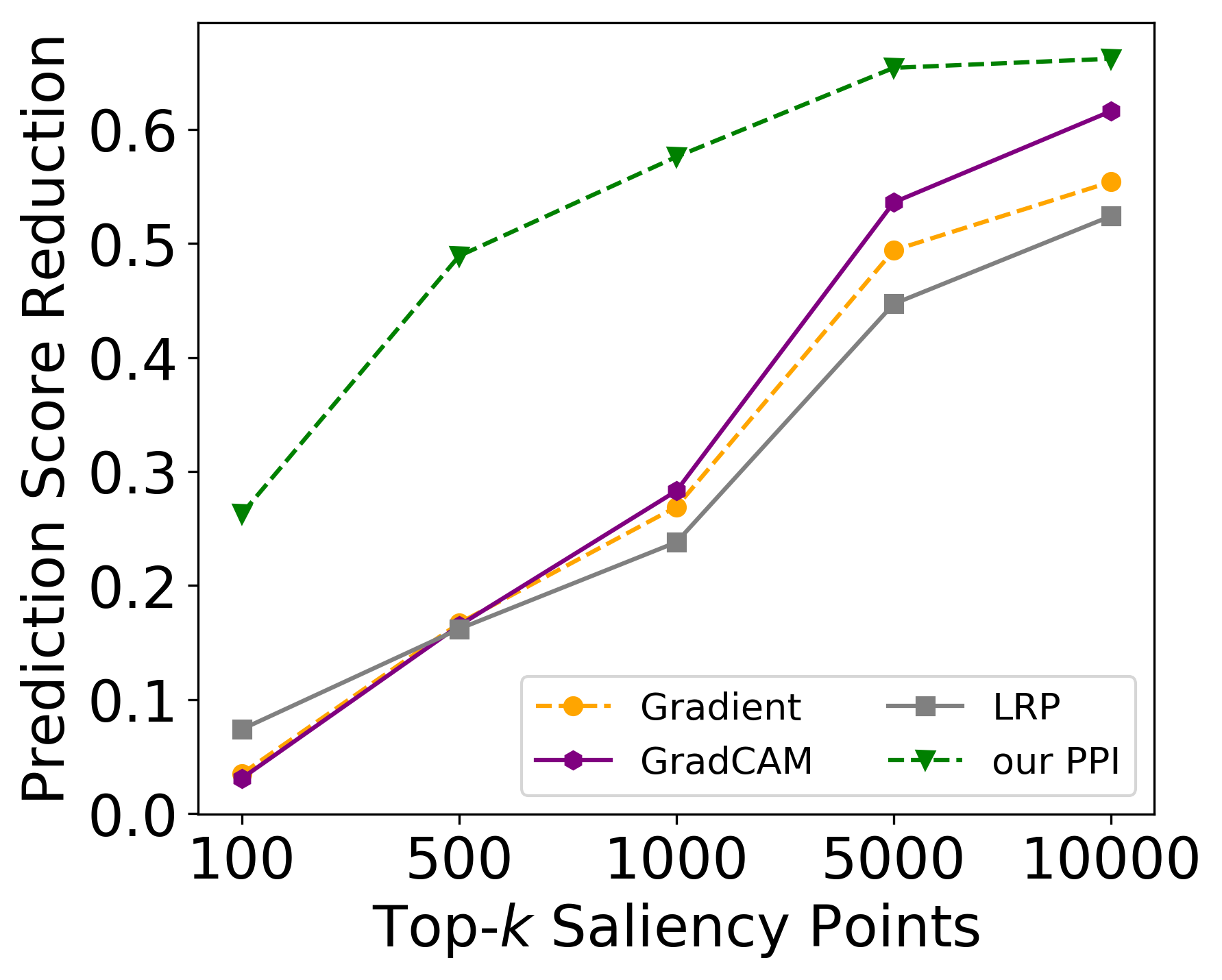}
        %  \vspace{-.2em}
         \caption{CUB}
         \label{fig:cub_point}
     \end{subfigure}
% \begin{subfigure}[b]{0.23\textwidth}
%          \centering
%          \includegraphics[width=\textwidth]{LaTeX/Figures/cub-energy.png}
%          \caption{Energy-based test}
%          \label{fig:cub_point_energy}
%      \end{subfigure}  
\begin{subfigure}[b]{0.23\textwidth}
         \centering
         \includegraphics[width=\textwidth]{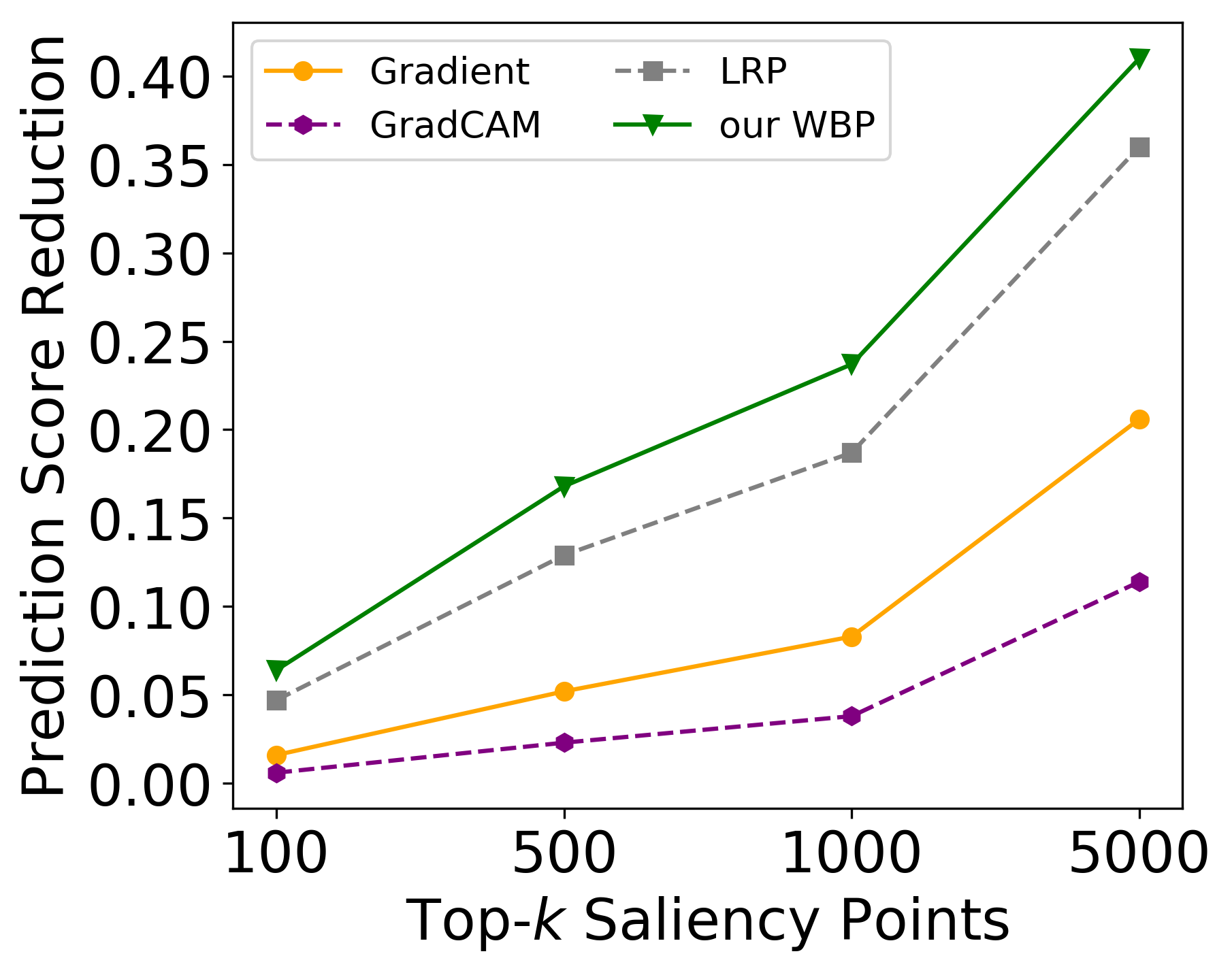}
        %  \vspace{-.2em}
         \caption{ImageNet}
         \label{fig:imagenet_point}
     \end{subfigure}
     \vspace{-.5em}
\caption{Quantitative evaluations of causal relevance of competing saliency maps (higher is better).} %on CUB and ImageNet 
\vspace{-1em}
\end{figure}

To quantitatively evaluate the causal relevance of competing saliency maps, we adopt the evaluation scheme proposed in \cite{hooker2019benchmark}, consisting of masking out the contributing saliency pixels and then calculating the reduction in prediction score.
A larger reduction is considered better for accurately capturing the pixels that `cause' the prediction.
Results are summarized in Figure~\ref{fig:cub_point}, where we progressively remove the top-$k$ saliency points, with $k=100, 500, 1000, 5000, 10000$ ($10000 \approx 6.6\%$ of all pixels), from the CUB test input images. %Table~\ref{tab:exp_ucb}
Our PPI consistently outperforms its counterparts, with its lead being most substantial in the low-$k$ regime.
Notably, for large $k$, PPI removes nearly all predictive signal.
This implies PPI specifically targets the causal features.
Quantitative evaluation with additional metrics are provided in the Appendix.

To test the performance of WBP itself (without being trained with PPI), we compare WBP with different approaches for a VGG11 model trained on ImageNet from PyTorch model zoo. Figure \ref{fig:exp_cub}(left) shows that saliency maps generated by WBP more concentrate on objects themselves. Also, thanks to the fine resolution of WBP, the model pays more attention to the patterns on the fur to identify the leopard (row 1). This is more visually consistent with human judgement. Figure~\ref{fig:imagenet_point} demonstrates WBP identifies more causal pixels on ImageNet validation images. 

% \begin{figure}[thpb!]
% \centering
% \includegraphics[width=0.22\textwidth]{./Figures/imagenet-point-game.png}
% \caption{Comparative  evaluations  of  pointing  game  and energy-based test on ImageNet (higher is better).}
% \label{fig:exp_imagenet}
% \vspace{-1em}
% \end{figure}

\subsection{OCT-GA: Geographic Atrophy Classification} \label{sec:exp_ga}
Next we show how the proposed PPI handles the challenges of small training data and heterogeneity in medical image datasets.
In this experiment (with our new dataset, that we will make public), each OCT volume image consists of $100$ scans of a $512 \times 1000$ sized image \cite{boyer2017pathophysiology}.
We use a multi-view CNN model ~\cite{su2015multi} to process such 3D OCT inputs, and use it as our baseline solution (see the Appendix for details).
We investigate how the different saliency mapping schemes ({\it i.e.}, Grad, GradCAM, LRP, WBP) work with PPI.
For WBP, we also tested the bounding box variant, denoted as WBP (box) (see the Appendix for details).
In Table \ref{tab:exp_ga}, we see consistent performance gains in AUC score via incorporating PPI training (from $0.877$ to $0.925$, can be improve to $0.937$ by PPI with WBP(box)), accompanied by the reductions in model variation evaluated by the standard deviations of AUC from the five-fold cross-validation.
The gains are most significant when using our WBP for saliency mapping.
We further compare the saliency maps generated by these different combinations.
We see that without the additional supervision from PPI, competing solutions like Grad, GradCAM and LRP sometimes yield non-sensible saliency maps (attending to image corners).
Overall, PPI encourages more concentrated and less noisy saliency maps.
Also, different PPI-based saliency maps agree with each other to a larger extent.
Our findings are also verified by experts (co-authors, who are ophthalmologists specializing in GA) confirming that the PPI-based saliency maps are clinically relevant by focusing on retinal layers likely to contain abnormalities or lesions.
% subject specialists, confirming the output PPI saliency maps make clinical sense via flagging the retinal layers that may contain abnormalities or lesions\footnote{On-going work prepared separately.}~\cite{kermany2018identifying}.
These results underscore the practical value of the proposed proactive interventions. 

% --------- Normal size table 3 ------- 
% \begin{table}[t!]
%     \caption{AUC results for GA prediction. We report means and standard deviations (SDs) from 5-fold cross-validation.}
%     \centering
%     \begin{tabular}{lll}
%         \toprule
%         \textbf{AUC} & \textbf{Mean}   & \textbf{STD}    \\ \midrule
%         % Random Forest~\cite{x} & 0.787 & 0.021\\ [5pt]%\hline
%         Classification & 0.877 & 0.040 \\  
%         +$\text{PPI}_{LRP}$ & {0.895} & {0.037} \\ 
%         +$\text{PPI}_{GradCAM}$ & {0.908} & {0.036} \\ 
%         +$\text{PPI}_{WBP}$ & {0.925} & {0.023} \\ 
%         +$\text{PPI}_{WBP (box)}$ & \textbf{0.937} & \textbf{0.015} \\ 
%         \bottomrule
%     \end{tabular}
%     \label{tab:exp_ga}
%     \vspace{-4mm}
% \end{table}

\noindent{\bf Cross-domain generalization.} Common to medical image applications is that training samples are usually integrated from a number of healthcare facilities ({\it i.e.}, domains), and that predictions are sometimes to be made on subjects at other facilities.
Despite big efforts to standardize the image collection protocols, with different imaging systems operated by technicians with varying skills, apparent domain shifts are likely to compromise the cross-domain performance of these models.
We show this phenomenon on the \texttt{GA} dataset in Table \ref{tab:exp_transform}, where source samples are collected from four different hospitals in different health systems (A, B, C and D, see the Appendix for details).
% The number in cell row $x$ column $y$ is 
Each cell contains the AUC of the model trained on site X (row) and tested on site Y (column), with same-site predictions made on hold-out samples.
A significant performance drop is observed for cross-domain predictions (off-diagonals) compared to in-domain predictions (diagonals).
With the application of PPI, the performance gaps between in-domain and cross-domain predictions are considerably reduced.
The overall accuracy gains of PPI further justify the utility of causally-inspired modeling.
Notably, site D manifests strong spurious correlation that help in-domain prediction but degrades out-of-site generalization, which is partly resolved by the proposed PPI. 

\begin{table}[]
\caption{AUC results for GA prediction with or without PPI. Models are trained on one site and cross-validated on the other sites. Darker color indicates better performance.}
\resizebox{\linewidth}{!}{
\begin{tabular}{@{}lllllll@{}}
\toprule
With $\text{PPI}$   & A     & B     & C    & D  & Mean & STD \\ \midrule
A  & \cellcolor{blue!40} 1.000 & \cellcolor{blue!25}0.906 & \cellcolor{blue!15}0.877 & \cellcolor{blue!15}0.865 & \textbf{0.912}	& \textbf{0.061}\\
B  & \cellcolor{blue!15}0.851 & \cellcolor{blue!35}0.975 & \cellcolor{blue!15}0.863 & \cellcolor{blue!25}0.910 & \textbf{0.900}	& 0.056\\
C & \cellcolor{blue!35}0.954 & \cellcolor{blue!15}0.875 & \cellcolor{blue!25}0.904 & \cellcolor{blue!25}0.931 & \textbf{0.916}	& \textbf{0.034}\\
D & \cellcolor{blue!5}0.824 & \cellcolor{blue!5}0.846 & \cellcolor{blue!15}0.853 & \cellcolor{blue!25}0.904 & \textbf{0.857}	& \textbf{0.034}\\ %\bottomrule
%\end{tabular}}
\hline
%\resizebox{\linewidth}{!}{
%\begin{tabular}{@{}lllllll@{}}
%\toprule
No PPI  & A     & B     & C    & D   & Mean & STD \\ \midrule
A  & \cellcolor{blue!40}1.000 & \cellcolor{blue!15}0.854 & \cellcolor{blue!5}0.832 & \cellcolor{blue!5} 0.827 & 0.878 & 0.082\\
B  & \cellcolor{blue!5}0.810 & \cellcolor{blue!15}0.874 & \cellcolor{blue!5}0.850 & \cellcolor{blue!25}0.906 & 0.860 & \textbf{0.040}\\
C & \cellcolor{blue!15}0.860 & 0.779 & \cellcolor{blue!15}0.873 & \cellcolor{blue!15}0.862  & 0.843 & 0.043\\
D & 0.748 & 0.792 & \cellcolor{blue!5}0.836 & \cellcolor{blue!35}0.961  & 0.834 & 0.092\\ \bottomrule
\end{tabular}}
\label{tab:exp_transform}
\vspace{-3mm}
\end{table}

\subsection{LIDC-IDRI: Lung Lesions Classification} \label{sec:exp_lidc}

\begin{wraptable}{r}{3.7cm}
  \vspace{-15pt}    % 对应高度1
    \caption{LIDC-IDRI classification AUC results. }
    \centering
    \resizebox{\linewidth}{!}{
    \begin{tabular}{ll}
        \toprule
        \textbf{Models} & \textbf{AUC}    \\ \midrule
        Tensor Net-X~\cite{efthymiou2019tensornetwork} & 0.823 \\
        DenseNet~\cite{huang2017densely} & 0.829 \\
        LoTeNet~\cite{selvan2020tensor} & 0.874 \\
        Inception\_v3~\cite{szegedy2016rethinking} & 0.921 \\  
        +$\text{PPI}_{GradCAM}$ & {0.933}  \\ 
        +$\text{PPI}_{Gradient}$ & {0.930}  \\ 
        +$\text{PPI}_{LRP}$ & {0.931}  \\ 
        +$\text{PPI}_{WBP}$ & {0.935}  \\ 
        +$\text{PPI}_{WBP (box)}$ & \textbf{0.941} \\ 
        \bottomrule
    \end{tabular}}
    \label{tab:exp_lidc}
    \vspace{-7pt} 
\end{wraptable}

To further examine the practical advantages of the proposed PPI in real-world applications, we benchmark its utility on LIDC-IDRI; a public lung CT scan dataset \cite{armato2011lung}.
We followed the preprocessing steps outlined in \cite{kohl2018probabilistic} to prepare the data, and adopted the experimental setup from \cite{selvan2020tensor} to predict lesions.
We use Inception\_v3~\cite{szegedy2016rethinking} as our base model for both standard classification and PPI-enhanced training with various saliency mapping schemes.
See the Appendix for details.
% ~\ref{appendix:lidc}

\vspace{2mm}
\noindent{\bf Lesion classification.} We first compare PPI to other specialized SOTA network architectures.
Table~\ref{tab:exp_lidc} summarizes AUC scores of Tensor Net-X~\cite{efthymiou2019tensornetwork}, DenseNet~\cite{huang2017densely}, LoTeNet~\cite{selvan2020tensor}, Inception\_v3~\cite{szegedy2016rethinking}, as well as our Inception\_v3 trained with and without $\text{PPI}_{WBP}$.
The proposed $\text{PPI}_{WBP (box)}$ leads the performance chart by a considerable margin, improving Inception\_v3 from $0.92$ to $0.94$.

\begin{figure}[t!]
\centering
\includegraphics[width=0.43\textwidth]{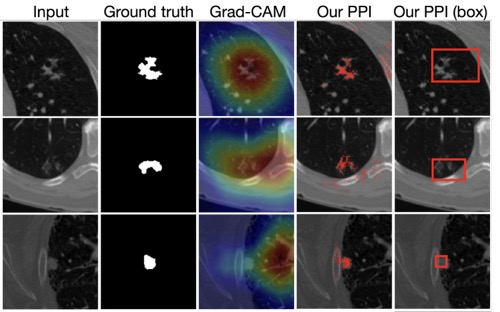}%exp_lidc.png
\caption{Saliency maps on LIDC-IDR. Saliency maps of PPI+WBP are mostly consistent with the ground truths.}
\label{fig:exp_lidc}
\vspace{-4mm}
\end{figure}

\vspace{2mm}
\noindent{\bf Weakly-supervised image segmentation.} In Figure~\ref{fig:exp_lidc}, we compare saliency maps generated by GradCAM, WBP, WBP (box) to the ground truth lesion masks from expert annotations.
Note that we have only supplied patch-label labels during training, not the pixel-level expert segmentation masks, which constitute a challenging task of weakly-supervised image segmentation.
In line with the observations from the GA experiment, our PPI-training enhanced WBP saliency maps are mostly consistent with the expert segmentations.
Together with Table \ref{tab:exp_lidc}, Figure \ref{fig:exp_lidc} confirms that the proposed PPI+WBP improves both the classification performance and model interpretability. 

\section{Conclusions}
We have presented \textit{Proactive Pseudo-Intervention} (PPI), a novel interpretable computer vision framework that organically integrates saliency mapping, causal reasoning, synthetic intervention and contrastive learning.
PPI couples saliency mapping with contrastive training by creating artificially intervened negative samples absent of causal features.
To communicate model insights and facilitate causal-informed reasoning, we derived an architecture-agnostic saliency mapping scheme called \textit{Weight Back Propagation} (WBP), which faithfully captures the causally-relevant pixels/features for model prediction.
Visual inspection of the saliency maps show that WBP, is more robust to spurious features compared to competing approaches.
Empirical results on natural and medical datasets verify the combination of PPI and WBP consistently delivers performance boosts across a wide range of tasks relative to competing solutions, and the gains are most significant where the application is complicated by small sample size, data heterogeneity, or confounded with spurious correlations.

%------------------------------------------------------------------------
% \section{Final copy}

% You must include your signed IEEE copyright release form when you submit
% your finished paper. We MUST have this form before your paper can be
% published in the proceedings.

{\small
\bibliographystyle{ieee_fullname}
\bibliography{main}

\begin{thebibliography}{10}\itemsep=-1pt

\bibitem{adebayo2018sanity}
Julius Adebayo, Justin Gilmer, Michael Muelly, Ian Goodfellow, Moritz Hardt,
  and Been Kim.
\newblock Sanity checks for saliency maps.
\newblock In {\em Advances in Neural Information Processing Systems}, pages
  9505--9515, 2018.

\bibitem{arjovsky2019invariant}
Martin Arjovsky, L{\'e}on Bottou, Ishaan Gulrajani, and David Lopez-Paz.
\newblock Invariant risk minimization.
\newblock {\em arXiv preprint arXiv:1907.02893}, 2019.

\bibitem{armato2011lung}
Samuel~G Armato~III, Geoffrey McLennan, Luc Bidaut, Michael~F McNitt-Gray,
  Charles~R Meyer, Anthony~P Reeves, Binsheng Zhao, Denise~R Aberle, Claudia~I
  Henschke, Eric~A Hoffman, et~al.
\newblock The lung image database consortium (lidc) and image database resource
  initiative (idri): a completed reference database of lung nodules on ct
  scans.
\newblock {\em Medical physics}, 38(2):915--931, 2011.

\bibitem{bach2015pixel}
Sebastian Bach, Alexander Binder, Gr{\'e}goire Montavon, Frederick Klauschen,
  Klaus-Robert M{\"u}ller, and Wojciech Samek.
\newblock On pixel-wise explanations for non-linear classifier decisions by
  layer-wise relevance propagation.
\newblock {\em PloS one}, 10(7):e0130140, 2015.

\bibitem{boyer2017pathophysiology}
David~S Boyer, Ursula Schmidt-Erfurth, Menno van Lookeren~Campagne, Erin~C
  Henry, and Christopher Brittain.
\newblock The pathophysiology of geographic atrophy secondary to age-related
  macular degeneration and the complement pathway as a therapeutic target.
\newblock {\em Retina (Philadelphia, Pa.)}, 37(5):819, 2017.

\bibitem{chang2018explaining}
Chun-Hao Chang, Elliot Creager, Anna Goldenberg, and David Duvenaud.
\newblock Explaining image classifiers by counterfactual generation.
\newblock In {\em International Conference on Learning Representations}, 2018.

\bibitem{chattopadhay2018grad}
Aditya Chattopadhay, Anirban Sarkar, Prantik Howlader, and Vineeth~N
  Balasubramanian.
\newblock Grad-cam++: Generalized gradient-based visual explanations for deep
  convolutional networks.
\newblock In {\em 2018 IEEE Winter Conference on Applications of Computer
  Vision (WACV)}, pages 839--847. IEEE, 2018.

\bibitem{chen2015deepdriving}
Chenyi Chen, Ari Seff, Alain Kornhauser, and Jianxiong Xiao.
\newblock Deepdriving: Learning affordance for direct perception in autonomous
  driving.
\newblock In {\em Proceedings of the IEEE International Conference on Computer
  Vision}, pages 2722--2730, 2015.

\bibitem{chen2020simple}
Ting Chen, Simon Kornblith, Mohammad Norouzi, and Geoffrey Hinton.
\newblock A simple framework for contrastive learning of visual
  representations.
\newblock {\em arXiv preprint arXiv:2002.05709}, 2020.

\bibitem{dabkowski2017real}
Piotr Dabkowski and Yarin Gal.
\newblock Real time image saliency for black box classifiers.
\newblock In {\em Advances in Neural Information Processing Systems}, pages
  6967--6976, 2017.

\bibitem{deng2009imagenet}
Jia Deng, Wei Dong, Richard Socher, Li-Jia Li, Kai Li, and Li Fei-Fei.
\newblock Imagenet: A large-scale hierarchical image database.
\newblock In {\em 2009 IEEE conference on computer vision and pattern
  recognition}, pages 248--255. Ieee, 2009.

\bibitem{deng2012mnist}
Li Deng.
\newblock The mnist database of handwritten digit images for machine learning
  research [best of the web].
\newblock {\em IEEE Signal Processing Magazine}, 29(6):141--142, 2012.

\bibitem{dhurandhar2018explanations}
Amit Dhurandhar, Pin-Yu Chen, Ronny Luss, Chun-Chen Tu, Paishun Ting,
  Karthikeyan Shanmugam, and Payel Das.
\newblock Explanations based on the missing: Towards contrastive explanations
  with pertinent negatives.
\newblock In {\em Advances in Neural Information Processing Systems}, pages
  592--603, 2018.

\bibitem{du2018towards}
Mengnan Du, Ninghao Liu, Qingquan Song, and Xia Hu.
\newblock Towards explanation of dnn-based prediction with guided feature
  inversion.
\newblock In {\em Proceedings of the 24th ACM SIGKDD International Conference
  on Knowledge Discovery \& Data Mining}, pages 1358--1367, 2018.

\bibitem{efthymiou2019tensornetwork}
Stavros Efthymiou, Jack Hidary, and Stefan Leichenauer.
\newblock Tensornetwork for machine learning.
\newblock {\em arXiv preprint arXiv:1906.06329}, 2019.

\bibitem{erhan2009visualizing}
Dumitru Erhan, Yoshua Bengio, Aaron Courville, and Pascal Vincent.
\newblock Visualizing higher-layer features of a deep network.
\newblock {\em University of Montreal}, 1341(3):1, 2009.

\bibitem{fong2019understanding}
Ruth Fong, Mandela Patrick, and Andrea Vedaldi.
\newblock Understanding deep networks via extremal perturbations and smooth
  masks.
\newblock In {\em Proceedings of the IEEE International Conference on Computer
  Vision}, pages 2950--2958, 2019.

\bibitem{fong2017interpretable}
Ruth~C Fong and Andrea Vedaldi.
\newblock Interpretable explanations of black boxes by meaningful perturbation.
\newblock In {\em Proceedings of the IEEE International Conference on Computer
  Vision}, pages 3429--3437, 2017.

\bibitem{fukui2019attention}
Hiroshi Fukui, Tsubasa Hirakawa, Takayoshi Yamashita, and Hironobu Fujiyoshi.
\newblock Attention branch network: Learning of attention mechanism for visual
  explanation.
\newblock In {\em Proceedings of the IEEE Conference on Computer Vision and
  Pattern Recognition}, pages 10705--10714, 2019.

\bibitem{girshick2014rich}
Ross Girshick, Jeff Donahue, Trevor Darrell, and Jitendra Malik.
\newblock Rich feature hierarchies for accurate object detection and semantic
  segmentation.
\newblock In {\em Proceedings of the IEEE conference on computer vision and
  pattern recognition}, pages 580--587, 2014.

\bibitem{goyal2019counterfactual}
Yash Goyal, Ziyan Wu, Jan Ernst, Dhruv Batra, Devi Parikh, and Stefan Lee.
\newblock Counterfactual visual explanations.
\newblock In {\em ICML}, 2019.

\bibitem{grill2020bootstrap}
Jean-Bastien Grill, Florian Strub, Florent Altch{\'e}, Corentin Tallec, Pierre
  Richemond, Elena Buchatskaya, Carl Doersch, Bernardo Avila~Pires, Zhaohan
  Guo, Mohammad Gheshlaghi~Azar, et~al.
\newblock Bootstrap your own latent-a new approach to self-supervised learning.
\newblock {\em Advances in Neural Information Processing Systems}, 33, 2020.

\bibitem{gutmann2010noise}
Michael Gutmann and Aapo Hyv{\"a}rinen.
\newblock Noise-contrastive estimation: A new estimation principle for
  unnormalized statistical models.
\newblock In {\em Proceedings of the Thirteenth International Conference on
  Artificial Intelligence and Statistics}, pages 297--304, 2010.

\bibitem{he2020momentum}
Kaiming He, Haoqi Fan, Yuxin Wu, Saining Xie, and Ross Girshick.
\newblock Momentum contrast for unsupervised visual representation learning.
\newblock In {\em Proceedings of the IEEE/CVF Conference on Computer Vision and
  Pattern Recognition}, pages 9729--9738, 2020.

\bibitem{he2016deep}
Kaiming He, Xiangyu Zhang, Shaoqing Ren, and Jian Sun.
\newblock Deep residual learning for image recognition.
\newblock In {\em Proceedings of the IEEE conference on computer vision and
  pattern recognition}, pages 770--778, 2016.

\bibitem{hooker2019benchmark}
Sara Hooker, Dumitru Erhan, Pieter-Jan Kindermans, and Been Kim.
\newblock A benchmark for interpretability methods in deep neural networks.
\newblock In {\em Advances in Neural Information Processing Systems}, pages
  9737--9748, 2019.

\bibitem{huang2017densely}
Gao Huang, Zhuang Liu, Laurens Van Der~Maaten, and Kilian~Q Weinberger.
\newblock Densely connected convolutional networks.
\newblock In {\em Proceedings of the IEEE conference on computer vision and
  pattern recognition}, pages 4700--4708, 2017.

\bibitem{johansson2016learning}
Fredrik Johansson, Uri Shalit, and David Sontag.
\newblock Learning representations for counterfactual inference.
\newblock In {\em International conference on machine learning}, pages
  3020--3029, 2016.

\bibitem{khosla2020supervised}
Prannay Khosla, Piotr Teterwak, Chen Wang, Aaron Sarna, Yonglong Tian, Phillip
  Isola, Aaron Maschinot, Ce Liu, and Dilip Krishnan.
\newblock Supervised contrastive learning.
\newblock {\em arXiv preprint arXiv:2004.11362}, 2020.

\bibitem{kohl2018probabilistic}
Simon Kohl, Bernardino Romera-Paredes, Clemens Meyer, Jeffrey De~Fauw, Joseph~R
  Ledsam, Klaus Maier-Hein, SM~Ali Eslami, Danilo~Jimenez Rezende, and Olaf
  Ronneberger.
\newblock A probabilistic u-net for segmentation of ambiguous images.
\newblock In {\em Advances in Neural Information Processing Systems}, pages
  6965--6975, 2018.

\bibitem{krizhevsky2009learning}
Alex Krizhevsky, Geoffrey Hinton, et~al.
\newblock Learning multiple layers of features from tiny images.
\newblock 2009.

\bibitem{krizhevsky2017imagenet}
Alex Krizhevsky, Ilya Sutskever, and Geoffrey~E Hinton.
\newblock Imagenet classification with deep convolutional neural networks.
\newblock {\em Communications of the ACM}, 60(6):84--90, 2017.

\bibitem{langlotz2019roadmap}
Curtis~P Langlotz, Bibb Allen, Bradley~J Erickson, Jayashree Kalpathy-Cramer,
  Keith Bigelow, Tessa~S Cook, Adam~E Flanders, Matthew~P Lungren, David~S
  Mendelson, Jeffrey~D Rudie, et~al.
\newblock A roadmap for foundational research on artificial intelligence in
  medical imaging: from the 2018 nih/rsna/acr/the academy workshop.
\newblock {\em Radiology}, 291(3):781--791, 2019.

\bibitem{leuschen2013spectral}
Jessica~N Leuschen, Stefanie~G Schuman, Katrina~P Winter, Michelle~N McCall,
  Wai~T Wong, Emily~Y Chew, Thomas Hwang, Sunil Srivastava, Neeru Sarin, Traci
  Clemons, et~al.
\newblock Spectral-domain optical coherence tomography characteristics of
  intermediate age-related macular degeneration.
\newblock {\em Ophthalmology}, 120(1):140--150, 2013.

\bibitem{li2018tell}
Kunpeng Li, Ziyan Wu, Kuan-Chuan Peng, Jan Ernst, and Yun Fu.
\newblock Tell me where to look: Guided attention inference network.
\newblock In {\em Proceedings of the IEEE Conference on Computer Vision and
  Pattern Recognition}, pages 9215--9223, 2018.

\bibitem{mahendran2016salient}
Aravindh Mahendran and Andrea Vedaldi.
\newblock Salient deconvolutional networks.
\newblock In {\em European Conference on Computer Vision}, pages 120--135.
  Springer, 2016.

\bibitem{montavon2019gradient}
Gr{\'e}goire Montavon.
\newblock Gradient-based vs. propagation-based explanations: an axiomatic
  comparison.
\newblock In {\em Explainable AI: Interpreting, Explaining and Visualizing Deep
  Learning}, pages 253--265. Springer, 2019.

\bibitem{montavon2019layer}
Gr{\'e}goire Montavon, Alexander Binder, Sebastian Lapuschkin, Wojciech Samek,
  and Klaus-Robert M{\"u}ller.
\newblock Layer-wise relevance propagation: an overview.
\newblock In {\em Explainable AI: interpreting, explaining and visualizing deep
  learning}, pages 193--209. Springer, 2019.

\bibitem{oord2018representation}
Aaron van~den Oord, Yazhe Li, and Oriol Vinyals.
\newblock Representation learning with contrastive predictive coding.
\newblock {\em arXiv preprint arXiv:1807.03748}, 2018.

\bibitem{o2020generative}
Matthew O'Shaughnessy, Gregory Canal, Marissa Connor, Christopher Rozell, and
  Mark Davenport.
\newblock Generative causal explanations of black-box classifiers.
\newblock {\em Advances in Neural Information Processing Systems}, 33, 2020.

\bibitem{pearl2009causality}
Judea Pearl.
\newblock {\em Causality}.
\newblock Cambridge university press, 2009.

\bibitem{petsiuk2018rise}
Vitali Petsiuk, Abir Das, and Kate Saenko.
\newblock Rise: Randomized input sampling for explanation of black-box models.
\newblock {\em arXiv preprint arXiv:1806.07421}, 2018.

\bibitem{rebuffi2020there}
Sylvestre-Alvise Rebuffi, Ruth Fong, Xu Ji, and Andrea Vedaldi.
\newblock There and back again: Revisiting backpropagation saliency methods.
\newblock In {\em Proceedings of the IEEE/CVF Conference on Computer Vision and
  Pattern Recognition}, pages 8839--8848, 2020.

\bibitem{ribeiro2016should}
Marco~Tulio Ribeiro, Sameer Singh, and Carlos Guestrin.
\newblock " why should i trust you?" explaining the predictions of any
  classifier.
\newblock In {\em Proceedings of the 22nd ACM SIGKDD international conference
  on knowledge discovery and data mining}, pages 1135--1144, 2016.

\bibitem{rieger2020interpretations}
Laura Rieger, Chandan Singh, William Murdoch, and Bin Yu.
\newblock Interpretations are useful: penalizing explanations to align neural
  networks with prior knowledge.
\newblock In {\em International Conference on Machine Learning}, pages
  8116--8126. PMLR, 2020.

\bibitem{ross2017right}
Andrew~Slavin Ross, Michael~C Hughes, and Finale Doshi-Velez.
\newblock Right for the right reasons: training differentiable models by
  constraining their explanations.
\newblock In {\em Proceedings of the 26th International Joint Conference on
  Artificial Intelligence}, pages 2662--2670, 2017.

\bibitem{ILSVRC15}
Olga Russakovsky, Jia Deng, Hao Su, Jonathan Krause, Sanjeev Satheesh, Sean Ma,
  Zhiheng Huang, Andrej Karpathy, Aditya Khosla, Michael Bernstein,
  Alexander~C. Berg, and Li Fei-Fei.
\newblock {ImageNet Large Scale Visual Recognition Challenge}.
\newblock {\em International Journal of Computer Vision (IJCV)},
  115(3):211--252, 2015.

\bibitem{sajda2006machine}
Paul Sajda.
\newblock Machine learning for detection and diagnosis of disease.
\newblock {\em Annu. Rev. Biomed. Eng.}, 8:537--565, 2006.

\bibitem{scholkopf2019causality}
Bernhard Sch{\"o}lkopf.
\newblock Causality for machine learning.
\newblock {\em arXiv preprint arXiv:1911.10500}, 2019.

\bibitem{selvan2020tensor}
Raghavendra Selvan and Erik~B Dam.
\newblock Tensor networks for medical image classification.
\newblock In {\em Medical Imaging with Deep Learning}, 2020.

\bibitem{selvaraju2017grad}
Ramprasaath~R Selvaraju, Michael Cogswell, Abhishek Das, Ramakrishna Vedantam,
  Devi Parikh, and Dhruv Batra.
\newblock Grad-cam: Visual explanations from deep networks via gradient-based
  localization.
\newblock In {\em Proceedings of the IEEE International Conference on Computer
  Vision}, pages 618--626, 2017.

\bibitem{seo2019regional}
Dasom Seo, Kanghan Oh, and Il-Seok Oh.
\newblock Regional multi-scale approach for visually pleasing explanations of
  deep neural networks.
\newblock {\em IEEE Access}, 8:8572--8582, 2019.

\bibitem{shrikumar2017learning}
Avanti Shrikumar, Peyton Greenside, and Anshul Kundaje.
\newblock Learning important features through propagating activation
  differences.
\newblock In {\em International Conference on Machine Learning}, pages
  3145--3153, 2017.

\bibitem{simonyan2013deep}
Karen Simonyan, Andrea Vedaldi, and Andrew Zisserman.
\newblock Deep inside convolutional networks: Visualising image classification
  models and saliency maps.
\newblock {\em arXiv preprint arXiv:1312.6034}, 2013.

\bibitem{simonyan2014very}
Karen Simonyan and Andrew Zisserman.
\newblock Very deep convolutional networks for large-scale image recognition.
\newblock {\em arXiv preprint arXiv:1409.1556}, 2014.

\bibitem{su2015multi}
Hang Su, Subhransu Maji, Evangelos Kalogerakis, and Erik Learned-Miller.
\newblock Multi-view convolutional neural networks for 3d shape recognition.
\newblock In {\em Proceedings of the IEEE international conference on computer
  vision}, pages 945--953, 2015.

\bibitem{suter2019robustly}
Raphael Suter, Djordje Miladinovic, Bernhard Sch{\"o}lkopf, and Stefan Bauer.
\newblock Robustly disentangled causal mechanisms: Validating deep
  representations for interventional robustness.
\newblock In {\em International Conference on Machine Learning}, pages
  6056--6065. PMLR, 2019.

\bibitem{szegedy2016rethinking}
Christian Szegedy, Vincent Vanhoucke, Sergey Ioffe, Jon Shlens, and Zbigniew
  Wojna.
\newblock Rethinking the inception architecture for computer vision.
\newblock In {\em Proceedings of the IEEE conference on computer vision and
  pattern recognition}, pages 2818--2826, 2016.

\bibitem{tian2019contrastive}
Yonglong Tian, Dilip Krishnan, and Phillip Isola.
\newblock Contrastive multiview coding.
\newblock {\em arXiv preprint arXiv:1906.05849}, 2019.

\bibitem{wagner2019interpretable}
Jorg Wagner, Jan~Mathias Kohler, Tobias Gindele, Leon Hetzel, Jakob~Thaddaus
  Wiedemer, and Sven Behnke.
\newblock Interpretable and fine-grained visual explanations for convolutional
  neural networks.
\newblock In {\em Proceedings of the IEEE Conference on Computer Vision and
  Pattern Recognition}, pages 9097--9107, 2019.

\bibitem{WahCUB_200_2011}
C. Wah, S. Branson, P. Welinder, P. Perona, and S. Belongie.
\newblock {The Caltech-UCSD Birds-200-2011 Dataset}.
\newblock Technical Report CNS-TR-2011-001, California Institute of Technology,
  2011.

\bibitem{wang2019learning}
Haohan Wang, Zexue He, Zachary~C Lipton, and Eric~P Xing.
\newblock Learning robust representations by projecting superficial statistics
  out.
\newblock {\em InInternational Conference on Learning Representations}, 2019.

\bibitem{wang2020score}
Haofan Wang, Zifan Wang, Mengnan Du, Fan Yang, Zijian Zhang, Sirui Ding, Piotr
  Mardziel, and Xia Hu.
\newblock Score-cam: Score-weighted visual explanations for convolutional
  neural networks.
\newblock In {\em Proceedings of the IEEE/CVF Conference on Computer Vision and
  Pattern Recognition Workshops}, pages 24--25, 2020.

\bibitem{wang2019sharpen}
Lezi Wang, Ziyan Wu, Srikrishna Karanam, Kuan-Chuan Peng, Rajat~Vikram Singh,
  Bo Liu, and Dimitris~N Metaxas.
\newblock Sharpen focus: Learning with attention separability and consistency.
\newblock In {\em Proceedings of the IEEE International Conference on Computer
  Vision}, pages 512--521, 2019.

\bibitem{wang2020visual}
Tan Wang, Jianqiang Huang, Hanwang Zhang, and Qianru Sun.
\newblock Visual commonsense representation learning via causal inference.
\newblock In {\em Proceedings of the IEEE/CVF Conference on Computer Vision and
  Pattern Recognition Workshops}, pages 378--379, 2020.

\bibitem{zeiler2014visualizing}
Matthew~D Zeiler and Rob Fergus.
\newblock Visualizing and understanding convolutional networks.
\newblock In {\em European conference on computer vision}, pages 818--833.
  Springer, 2014.

\bibitem{zhang2020causal}
Cheng Zhang, Kun Zhang, and Yingzhen Li.
\newblock A causal view on robustness of neural networks.
\newblock {\em arXiv preprint arXiv:2005.01095}, 2020.

\bibitem{zhao2018uniqueness}
Yitian Zhao, Yalin Zheng, Yifan Zhao, Yonghuai Liu, Zhili Chen, Peng Liu, and
  Jiang Liu.
\newblock Uniqueness-driven saliency analysis for automated lesion detection
  with applications to retinal diseases.
\newblock In {\em International Conference on Medical Image Computing and
  Computer-Assisted Intervention}, pages 109--118. Springer, 2018.

\bibitem{zhou2016learning}
Bolei Zhou, Aditya Khosla, Agata Lapedriza, Aude Oliva, and Antonio Torralba.
\newblock Learning deep features for discriminative localization.
\newblock In {\em Proceedings of the IEEE conference on computer vision and
  pattern recognition}, pages 2921--2929, 2016.

\end{thebibliography}
}

\newpage
\clearpage

\appendix
\begin{figure*}[t!]
    \centering
    \includegraphics[width=0.95\textwidth]{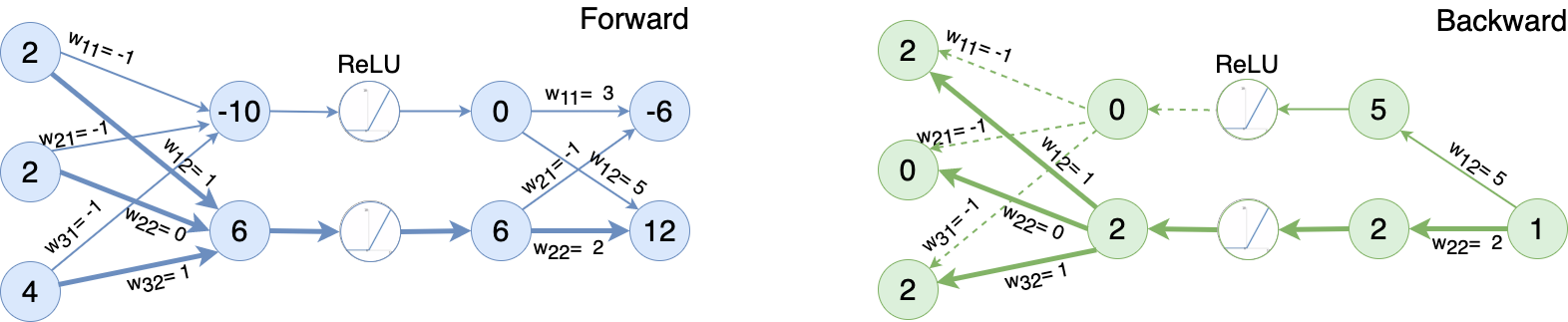}
    \caption{Illustration of the Weight Backpropagation (WBP) through a fully connected layer with ReLU activation layer.}
    \label{fig:wb_pool}
    \vspace{-1em}
 \end{figure*}
 
\section{Weight Backpropagation (WBP)} 
\subsection{Graphical illustration of WBP}
See Figure~\ref{fig:wb_pool} for a graphical illustration with standard affine mapping and ReLU activation.

\subsection{Derivation of Convolutional Weight Backpropagation}\label{appendix:cwb}
Let's denote the input variable as $I\in{\mathbb{R}^{H\times{W}}}$, the convolutional filter weight as $W\in{\mathbb{R}^{(2S+1)\times{(2S+1)}}}$, the output variable as $O\in{\mathbb{R}^{H\times{W}}}$, and the weight backpropagate to $O$ as $\hat{W}\in{\mathbb{R}^{H\times{W}}}$. We omit the  bias here because it does not directly interact with the input variables. We denote $\otimes$ as the convolutional operator. We have
\resizebox{.5\textwidth}{!}
{
\begin{minipage}{\linewidth}
\begin{align*}
    O &= I\otimes{W} \\
    O_{i,j} &= \sum_{i'=-S}^{S}\sum_{j'=-S}^{S}I_{i+i',j+j'}W_{i'+S,j'+S} \\
    \sum_i\sum_jO_{i,j}\tilde{W}_{i,j} &= \sum_i\sum_j\sum_{i'=-S}^{S}\sum_{j'=-S}^{S}I_{i+i',j+j'}W_{i'+S,j'+S}\tilde{W}_{i,j} \\
    \sum_i\sum_jO_{i,j}\tilde{W}_{i,j} &= \sum_i\sum_jI_{i,j}\sum_{i'=-S}^{S}\sum_{j'=-S}^{S}\tilde{W}_{i+i',j+j'}W_{-i'+S,-j'+S} \\
    \sum_i\sum_jO_{i,j}\tilde{W}_{i,j} &= \sum_i\sum_jI_{i,j}(\tilde{W}\otimes{[W]_{flip_{i,j}}})_{i,j}
\end{align*}
\end{minipage}
}

Hence the weight backpropagate through a convolutional layer is $\tilde{W}^l=\tilde{W}^{l+1}\otimes{[W^l]_{flip}}$. For the 3D cases, $I^l\in{\mathbb{R}^{D_1\times{H}\times{W}}}$,the weight back propagates to $O^l$ is $\tilde{W}^{l+1}\in{\mathbb{R}^{D_2\times{H}\times{W}}}$ and the convolutional weight is $W^l\in{\mathbb{R}^{D_2\times{D_1}\times{(2S+1)}\times{(2S+1)}}}$. To match the depth of $\tilde{W}^{l+1}$, the $W^l$ is transposed in the first two dimensions. So 
$\tilde{W}^{l}=\tilde{W}^{l+1}\otimes{[W^{l}]^{T_{0,1}}_{flip_{2,3}}}$. If the convolutional layer is downsizing the input variable ({\it i.e.}, strides), the $\tilde{W}^{l+1}_{ijk}$ is  padded with zeros around the weights (left,right,up, and down) to for the input elements that the convolutional filter strides over. The number of padding zeros is equal to the number of strides minus 1.

% \section{Pooling}
% \begin{figure*}[htp]
%      \centering
%      \begin{subfigure}[b]{0.35\textwidth}
%          \centering
%          \includegraphics[width=\textwidth]{Figures/avg pool.png}
%          \caption{Average Pool}
%          \label{fig:avgpool}
%      \end{subfigure}
%     %  \hfill
%      \begin{subfigure}[b]{0.35\textwidth}
%          \centering
%          \includegraphics[width=\textwidth]{Figures/max pool.png}
%          \caption{Max Pool}
%          \label{fig:maxpool}
%      \end{subfigure}
%      \caption{Illustration of the weight backpropagation through pooling layers.}
%      \label{fig:wb_pool}
%  \end{figure*}
% \begin{figure}[htpb]
% \centering
% \includegraphics[width=0.45\textwidth]{./Figures/examples_intro_ga_2.jpg}
% \caption{Attention maps generation byGrad-CAM, and our Weight Back Propagation (WBP).}
% \label{fig:examples_into_ga}
% \end{figure}

\section{Details on Causal Masking} \label{appedix:transformation}
In this work, we consider three types of causal masking: ($i$) the point-wise soft causal masking defined by Equation (2) in the main text, ($ii$) hard masking, and ($iii$) box masking. For the hard masking, for each image, we keep points with WBP weight larger than $k$ times of the standard deviation of WBP weights of the whole image. We test $k$ from $1$ to $7$ and achieve similar results. As $k=7$ performs slightly better, we set $k$ as $7$ for all experiments. For the box masking, we use the center of mass for these kept points as the center to draw a box. The height and width of this box is defined as $center_{h/w}\pm{1.2std_{h/w}}$. In this way at least $90\%$ of filtered points are contained in the box. For the soft masking, we set $\omega$ to $100$ and $\sigma$ to $0.25$. We have also experimented with image-adaptive thresholds instead of a fixed $\sigma$ for all inputs, {\it i.e.}, set the threshold as mean value plus $k$ times of the standard deviation of WBP weights of the whole image. We repeat the experiments a few times and the results are consistent. The experiment comparison of these masking methods mention above is conducted on LIDC dataset.  %\tao{what's the value of $k$?}

\begin{table}[htb!]
    \caption{Different causal masking methods on LIDC}
    \centering
    \begin{tabular}{ll}
        \toprule
        \textbf{Models} & \textbf{AUC}  \\ \midrule
        WBP-soft (fixed $\sigma$) & 0.931 \\  
        WBP-soft (adaptive $\sigma$) & {0.941} \\ 
        WBP-hard (point) & {0.935} \\ 
        WBP-hard (box) & \textbf{0.941} \\ 
        \bottomrule
    \end{tabular}
    \label{tab:mask_lidc}
\end{table}

\section{Related Work} 
In this work, we propose a contrastive causal representation learning
strategy, i.e., Proactive Pseudo-Intervention (PPI), that leverages proactive interventions to identify causally-relevant image features.  This approach is complemented with a novel causal salience map visualization module, i.e., Weight Back Propagation (WBP), that identifies important pixels in the raw input image, which greatly facilitates interpretability of predictions.

Prior related works will be discussed in this section. Compared with alternative post-hot saliency mapping methods, WBP outperforms these methods as both a standalone causal saliency map and a trainable model for model interpretation. Compared with other trainable interpretation models, the proposed PPI+WBP improves both model performance and model interpretations.

\subsection{Post-hoc Saliency Maps}
We compare WBP with other post-hoc saliency mapping methods to show why WBP is  able to target the causal features, and generate more succinct and reliable saliency maps. 

\textbf{Perturbation Based Methods}
These methods make perturbations to individual inputs or neurons and monitor the impact on output neurons in the network. 
\cite{zeiler2014visualizing} occludes different segments of an input image and visualized the change in the activations of subsequent layers. Several methods follow a similar idea, but use other importance measures or occlusion strategies \cite{petsiuk2018rise, ribeiro2016should, seo2019regional}.
More complicated works aim to generate an explanation by optimizing for a perturbed version of the image \cite{fong2017interpretable, fong2019understanding, dabkowski2017real, du2018towards}. \cite{wagner2019interpretable} proposes a new adversarial defense technique which filters gradients during optimization to achieve fine-grained explanation.
However, such perturbation based methods are computationally intensive and involve sophisticated model designs, which make it extremely hard to be integrated with other advance learning strategies.
%Interpretable and Fine-Grained Visual Explanations for Convolutional Neural Networks (adversarial defense) \cite{wagner2019interpretable} 

\textbf{Backpropagation Based Methods}
Backpropagation based methods (BBM) propagate an importance signal from an output neuron backwards through the layers to the input. These methods are usually fast to compute and produce fine-grained importance/relevancy maps. WBP is one of such method. 

The pioneer methods in this category backpropagate a gradient to the image, and branches of studies extend this work by manipulating the gradient. These methods are discussed and compared in~\cite{mahendran2016salient, erhan2009visualizing}. %The visualizations are fine-grained but not class-specific, where visualizations for different classes are nearly identical.
However, these maps are generally less concentrated~\cite{dabkowski2017real,fong2017interpretable} and less interpretable. 
Other BBMs such as Layer-wise Relevance Propagation~\cite{bach2015pixel}, DeepLift~\cite{shrikumar2017learning} employ top-down relevancy propagation rules. {DeepLift} is sensitive to the reference inputs, which needs more human efforts and background knowledge to produce appealing saliency maps. The nature of depending on reference inputs limits its ability on model diagnosis and couple with learning strategies to continuously improving models' performance.
%{LRP} We supply detail comparisons between WBP and variations of the state-of-the-art saliency map generation method, {\it i.e.,} LRP. 
LRP decomposes the relevance, $R$, from a neuron, $k$, in the upper layer to every connected neurons, $j$, in the lower layer. The decomposition is distributed through gradients under the suggested implementation \cite{montavon2019layer}. Our experiments on GA and CUB datasets show that vanilla LRP performs similar to gradient based methods, which is also demonstrated in \cite{montavon2019gradient}. The variants of LRP use complex rules to prioritize positive or large relevance, making the saliency map visually appealing to human. However, our experiments demonstrate the unfaithfulness of LRP and its variants as they highlight spuriously correlated features (boarderlines and backgrounds). By contrast, our WBP backpropagates the the weights of through layers to compute the contributions of each input pixel, which is truly faithful to the model, and WBP tends to highlight the target objects themselves rather than the background. At the same time, the simplicity and efficiency makes WBP easily work with other advanced learning strategies for both model diagnosis and improvements during training. % shall we mention the LRP is easily affected by the gradient and sometimes focus on strange areas, as shown in GA examples before PPI?

\label{ap:lrp}
\begin{table}[htp]
\caption{A list of commonly used LRP rules.\cite{montavon2019layer}}
    \centering
        \begin{tabular}{l | l}
        \toprule
        Rules & Formula\\ \midrule
        $LRP$ & $R_j=\sum_k\frac{a_jw_{jk}}{\sum_{0,j}a_jw_{jk}}R_k$\\ \hline
        $LRP_{\epsilon}$ & $R_j=\sum_k\frac{a_jw_{jk}}{\epsilon+\sum_{0,j}a_jw_{jk}}R_k$\\ \hline
        $LRP_{\gamma}$ & $R_j=\sum_k\frac{a_j(w_{jk}+\gamma w^{+}_{jk})}{\sum_{0,j}a_j(w_{jk}+\gamma w^{+}_{jk})}R_k$\\ \hline
        $LRP_{\alpha\beta}$  & $R_j=\sum_k(\alpha\frac{(a_jw_{jk})^+}{\sum_{0,j}(a_jw_{jk})^+}-\beta\frac{(a_jw_{jk})^-}{\sum_{0,j}(a_jw_{jk})^-})R_k$ \\ \hline
        $LRP_{flat}$  & $R_j=\sum_k\frac{1}{\sum_{j}1}R_k$\\ \hline
        $LRP_{w^2}$  &  $R_j=\sum_j\frac{w_{ij}^2}{\sum_iw^2_{ij}}R_j$\\ \hline
        $LRP_{Z^\beta}$  & $R_j=\sum_j\frac{x_iw_{ij}-l_iw^+_{ij}-h_iw^-{ij}}{\sum_ix_iw_{ij}-l_iw^+_{ij}-h_iw^-{ij}}R_j$\\ \bottomrule
        \end{tabular}
        
        \label{tbl:lrp_rules}
    \end{table}

\textbf{Activation Based Methods} Methods under this category (such as CAM, Grad-CAM, guided Grad-CAM, Grad-CAM++) use a linear combination of class activation maps from convolutional layers to derive a saliency map. The main difference between them is how to the linear combination weights are computed. %CAM generates class activation maps highlighting task-relevant regions by replacing fully-connected layers with convolution and global average pooling. Grad-CAM solves CAM's inflexibility where without changing the model architecture and retraining the parameters, class-wise attention maps were generated by means of gradients of the final prediction w.r.t. pixels in feature maps. 
% These methods generate easy to interpret heat-maps which can be overlaid on the image.
The generation of saliency maps is easy and these methods can be coupled with advanced training strategies to improve training \cite{li2018tell}. However, they fail at visualizing fine-grained evidence, which is particularly important in explaining medical classification models. Additionally, it is not guaranteed that the resulting explanations are faithful and reflect the decision making process of the model~\cite{du2018towards, selvaraju2017grad, wagner2019interpretable}. Grad-CAM++~\cite{chattopadhay2018grad} proposes to introduce higher-order derivatives to capture pixel-level importance, while its high computational cost in calculating the second- and third-order derivatives makes it impractical for training purposes.

\begin{figure*}[thpb]
\centering
\includegraphics[width=0.99\textwidth,trim=0 0 0 0, clip]{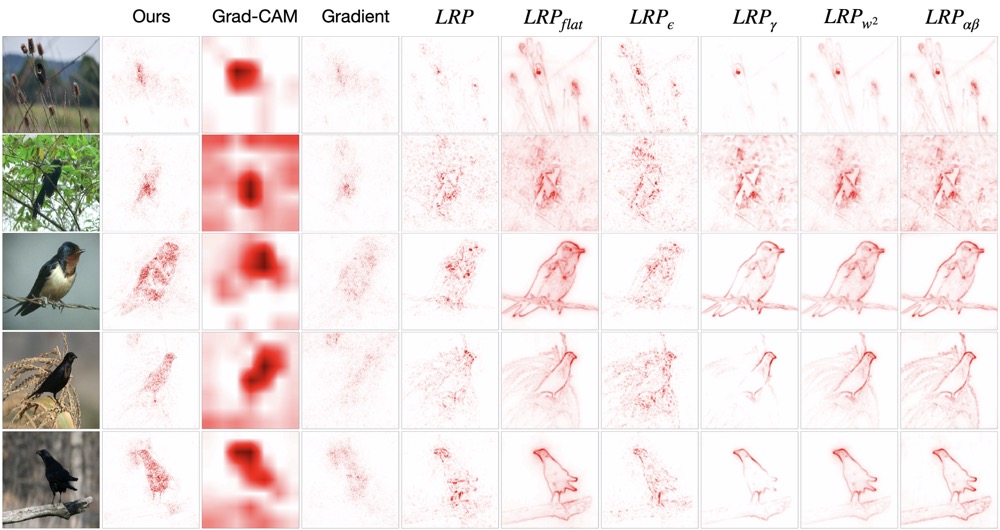} %35mm
\caption{Saliency maps on CUB dataset.}
\label{fig:cub_appendix}
\end{figure*}

\subsection{Interpretable Models}
Unlike the {\em post hoc} saliency map generation described above, an alternative approach is to train a separate module to explicitly produce model explanations \cite{fukui2019attention, goyal2019counterfactual, chang2018explaining,  fong2017interpretable, shrikumar2017learning}.
Such {\it post hoc} causal explanations can be generated with black-box classifiers based on a learned low-dimensional representation of the data \cite{o2020generative} .
Related to our work is adversarial-based visual explanation method is developed in \cite{wagner2019interpretable}, highlighting the key features in the input image for a specific prediction. 
Contrastive explanations are produced in \cite{dhurandhar2018explanations} to justify the predictions from a deep neural network.
Also in \cite{goyal2019counterfactual} the authors generate counterfactual visual explanations that highlight what and how regions of an image would need to change in order for the model to predict a {\em distractor} class $c'$ instead of the predicted class $c$.
The main differences to our construction are two fold: ($i$) they rely on a separate module to be trained, and ($ii$) these approaches only produce explanations, but such explanations are not exploited to provide feedback for model improvement.

Striking the goal of both good explanation and good performance is more challenging. 
% The change in classifier outputs is considered in 
One promising direction is to inject model-dependent perturbations to the input images as strategic augmentations
\cite{fong2017interpretable, dabkowski2017real, chang2018explaining}. In such examples, parts of the image have been masked and replaced with various references such as mean pixel values, blurred image regions, random noise, outputs of generative models, {\it etc}.
However, these pixel-level perturbations are very costly and difficult to craft.
%Sharpen Focus: Learning with Attention Separability and Consistency 
\cite{wang2019sharpen} propose new learning objectives for attention separability and cross-layer consistency, which result in improved attention discriminability and reduced visual confusion. However, it generates heat-map style attention maps, which fail in fine-resolution model explanations which is important in medical related tasks.  
In \cite{fukui2019attention} an additional attention branch is learned to generate attention map, and then applies the attention map to the original image or feature map;
they achieve compelling attention maps on natural images.
However, as the attention maps are not derived directly from the classification model, there is no guarantee for their faithfulness.
Further, having an additional attention network results in increased network size,
% sometimes doubling the size of the model, 
which raises concerns for the risk of over-fitting, particularly on datasets with a limited sample size.

\begin{figure*}[thpb]
\centering
\includegraphics[width=0.99\textwidth]{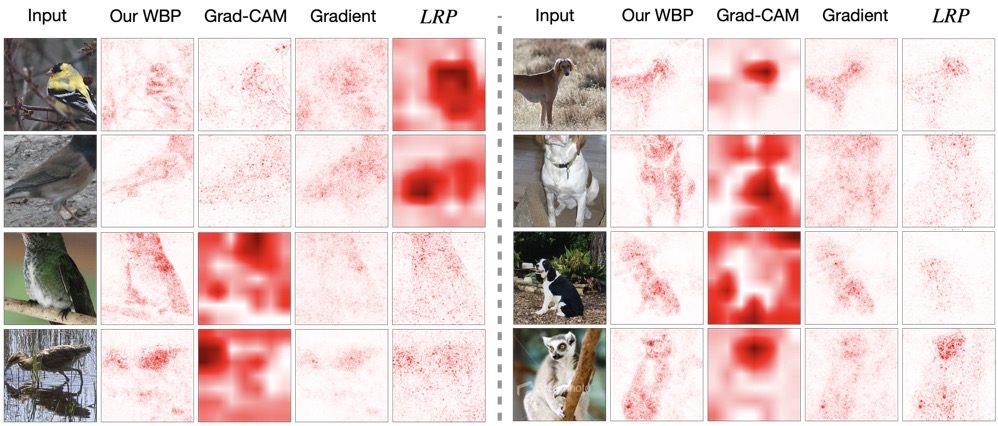}
\caption{Saliency maps on ImageNet dataset.}
\label{fig:example_imagenet}
\end{figure*}

\section{Natural Image Datasets}
\subsection{CUB Experiments}
\textbf{CUB dataset descriptions and experiment settings.}
CUB has 11,788 images of 200 bird spices. To train a VGG11 network, we use 8,190 training images and validate the model on 2,311 validation images, with the accuracy are reported on the remaining 1,227 testing images. The network is trained for 100 epochs with a learning rate decay of 0.1 every 30 epochs. The batch size is 32. The optimizer is SGD with initial learning rate at 0.01.  

\textbf{Classification performance improvement with PPI.}
We compare classification performance among models trained with different schemes. 
The baseline is VGG11 classification without PPI. Three different saliency mapping methods are tested within our PPI framework: Gradient, LRP, Grad-CAM, and WBP. Top 1k points with the largest saliency scores in all saliency maps are used to generate the soft mask so the comparison is fair.
During training, since only a small portion of points are used to generate the mask, the scale of $L_{con}$ is about 100 times smaller than other losses. To fix this imbalance, $L_{con}$ is weighted by a factor of 100$\times$ after the first 20 epochs. The results are shown in Table 2. % \ref{tab:exp_ga}.%{tab:cub_acc}.

% \begin{table}[htb]
%     \caption{Accuracy on CUB}
%     \centering
%     \begin{tabular}{ll}
%         \toprule
%         \textbf{Models} & \textbf{Accuracy}  \\ \midrule
%         % Random Forest~\cite{x} & 0.787 & 0.021\\ [5pt]%\hline
%         VGG 11 & 0.662 \\  
%         +$\text{PPI}_{LRP}$ & {0.680} \\ 
%         +$\text{PPI}_{Grad-CAM}$ & {0.683} \\ 
%         +$\text{PPI}_{WBP}$ & \textbf{0.696} \\ 
%         \bottomrule
%     \end{tabular}
%     \label{tab:cub_acc}
% \end{table}

\textbf{Model Interpretability} More saliency maps on CUB dataset are shown in Figure.~\ref{fig:cub_appendix}. Visually, gradient-based solutions (Grad and GradCAM) tend to yield overly dispersed maps, indicating a lack of specificity.
LRP gives more appealing saliency maps. However, these maps also heavily attend to the spurious background cues that presumably help with predictions. When trained with PPI, the saliency maps attend to birds body, and with WBP, the saliency maps focus on the causal related pixels. \textbf{Quantitative Test.} 
We also measure the quality of the generated saliency maps through their localization ability. Extending from pointing game which uses only the maximum point (energy-based pointing game \cite{wang2020score}), we care about how much energy of the saliency map falls into the target object bounding box. 
Evaluation result is shown in Table~\ref{tab:cub_energy}. We see our WBP outperforms prior arts by a large margin, more than $93\%$ energy of the top $100$ saliency points fall into the ground truth bounding box of the target object.

\begin{table*}[htpb]
\caption{Energry based test on CUB dataset (higher is better).}
\centering
\begin{tabular}{llllllllll}
    \toprule
    Point Num  & LRP   & LRP$_{flat}$
    & LRP$_{\epsilon}$ 
    & LRP$_{\gamma}$ & LRP$_{w^2}$    
    & LRP$_{\alpha\beta}$ & Gradient & GradCAM & WBP   \\ 
    \midrule
    100        & 0.819 & 0.671 & 0.819   & 0.685 & 0.684 & 0.588     & 0.796    & 0.730     & \textbf{0.935} \\
    500        & 0.772 & 0.639 & 0.772   & 0.632 & 0.644 & 0.542     & 0.744    & 0.731     & \textbf{0.890} \\
    1000       & 0.740 & 0.622 & 0.740   & 0.607 & 0.624 & 0.517     & 0.706    & 0.723     & \textbf{0.855} \\
    5000       & 0.624 & 0.575 & 0.624   & 0.549 & 0.579 & 0.449     & 0.568    & 0.637     & \textbf{0.720} \\
    10000      & 0.555 & 0.542 & 0.555   & 0.519 & 0.554 & 0.418     & 0.489    & 0.553     & \textbf{0.642} \\
    All points & 0.396 & 0.339 & 0.396   & 0.401 & 0.399 & 0.262     & 0.331    & 0.275     & \textbf{0.484} \\
    \bottomrule
\end{tabular}
\label{tab:cub_energy}
\end{table*}

\subsection{ImageNet Experiments}
To verify the effectiveness of WBP to find the relevance pixels on a large scale natural images, we generate saliency maps using an ImageNet pretrained VGG11 on ImageNet 2012 Validation set. Training is not part of this experiment. The saliency mapping methods evaluated with ImageNet are Gradient, Grad-CAM, LRP, and WBP. More saliency maps on ImageNet dataset are shown in Figure.~ \ref{fig:example_imagenet}. 
In the pointing game experiment, for each saliency method, we remove the top k saliency pixels and measure the drop in prediction scores are measured. Intuitively, a larger drop is considered a stronger evidence that the identified features are causal.

\subsection{CIFAR-10 Experiments}
To verify the effectiveness of PPI on improving the classification performance, we evaluate PPI with 4 different saliency mapping methods, namely Gradient, Grad-CAM, LRP, and WBP, during training. For all training including without PPI, all loss terms are weighted such that they are at a similar magnitude. The initial learning rate is 0.01 and reduce by $1/10$ every 30 epochs and we train the model for 100 epochs. The backbone is VGG11 with random initialized weights. 

\section{Geographic Atrophy (GA) Experiments}
\begin{figure*}[htpb]
\centering
\includegraphics[width=0.8\textwidth]{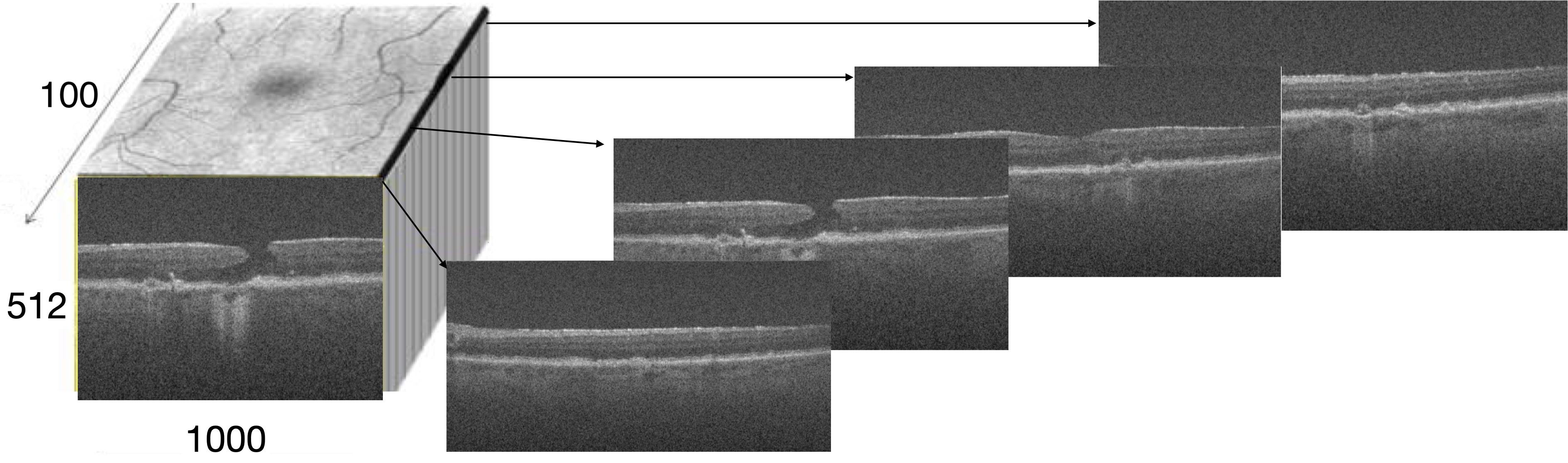}
\caption{Illustration of a 3D OCT image example.}
\label{fig:example_oct}
\end{figure*}

\textbf{GA dataset descriptions}
Our GA dataset is derived from the A2A SD-OCT Study (\url{http://ClinicalTrials.gov} identifier NCT00734487), which was an ancillary observational prospective study of a subset of eyes from the AREDS2 conducted at four sites (National Eye Institute, Duke Eye Center, Emory Eye Center, and Devers Eye Institute)~\cite{leuschen2013spectral}. 
In this experiment (with our new dataset, that we will make public), each OCT volume image consists of $100$ scans, each of which being a $512 \times 1000$ pixel image \cite{boyer2017pathophysiology}. $1,085$ OCT images are collected from $275$ subjects during $5$ years. An example of 3D OCT images is shown in Figure.~\ref{fig:example_oct}.

\subsection{Image differences between $4$ sites}
\begin{figure*}[htpb]
\centering
\includegraphics[width=0.8\textwidth]{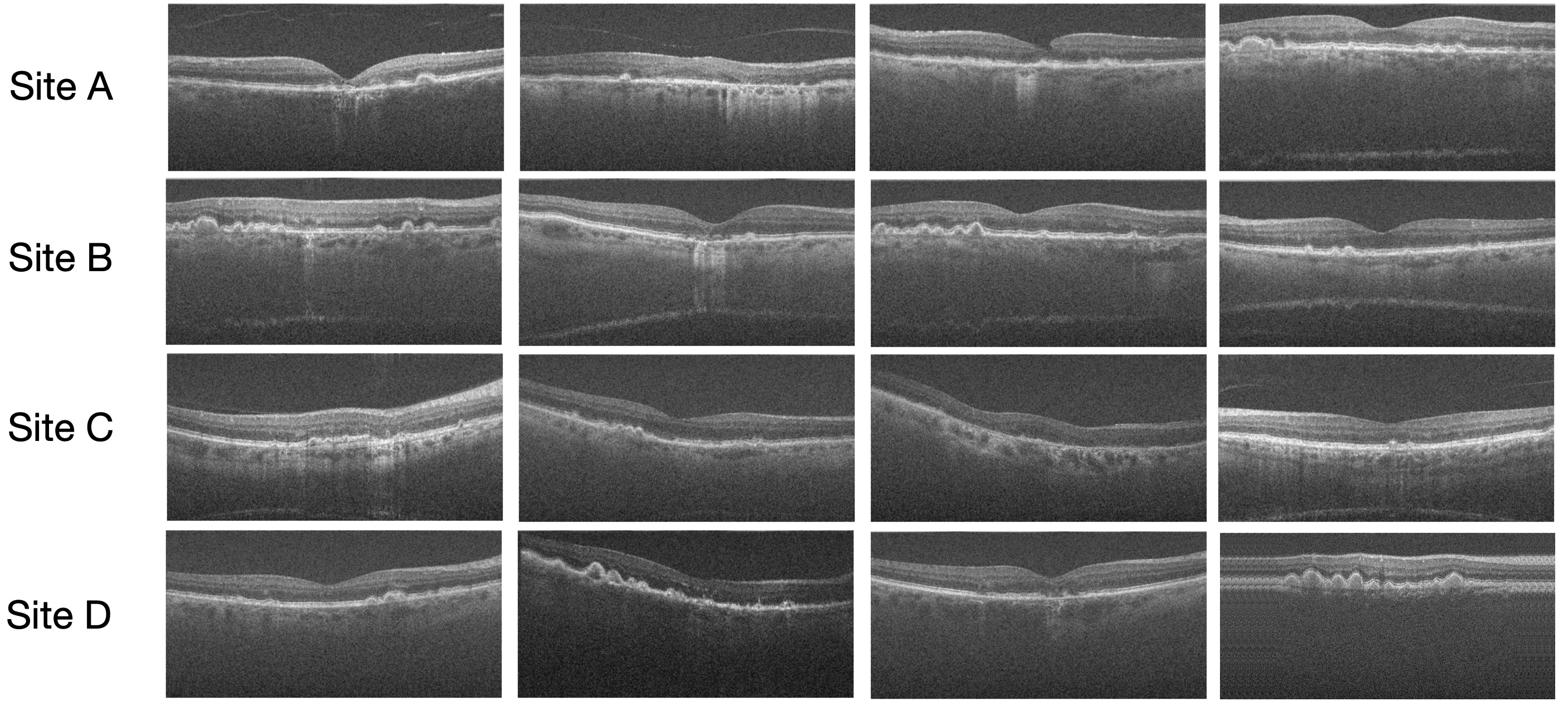}
\caption{OCT slice examples from $4$ site.}
\label{fig:example_4sites}
\end{figure*}

Images in GA dataset are collected from $4$ different sites, hereafter denoted as A, B, C, and D respectively. There are $315$ images ($101$ positive samples) from site A, $334$ images  ($73$ positive samples) from site B, $260$ images  ($131$ positive samples) from site A, and $176$ images  ($59$ positive samples) from site D. 
We show typical example images from $4$ sites separately in Figure.~\ref{fig:example_4sites}. As the dataset is collected during $7$ years, some images in site D are of smaller image size as they are sampled with different type of machine. We paddle these images by repeating the left and right areas, as show in the right bottom example. 

% (add image comparisons, show why site D is different from others.)

\subsection{Multi-view CNN Variation}\label{sec:multiview}
\begin{figure*}[htpb]
\centering
\includegraphics[width=0.9\textwidth]{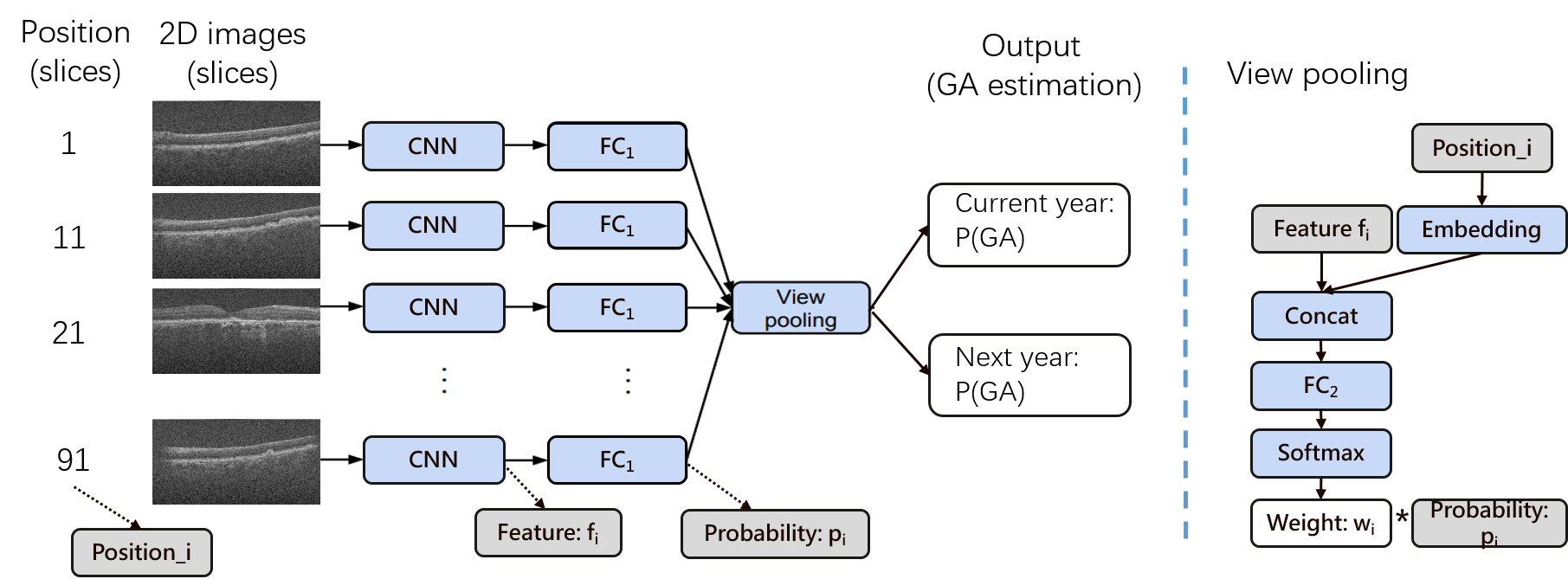}
\caption{Illustration of multi-view CNN based 3D OCT image classification model.}
\label{fig:multiview_cnn}
\end{figure*}

We use a variant of the multi-view CNN model ~\cite{su2015multi} to process the 3D OCT inputs, and use it as our baseline solution. The architecture of this model is outlined in Figure \ref{fig:multiview_cnn}. %This model uses inception\_v3 model as backbone model to generate GA scores for each 2D slice and combines these scores with our special designed slice-position-aware attention view pooling. 
For each slice, the model feed it into a CNN network, and get the feature $f_i$ of slice $i$ ($f_i = CNN( x_i ) $), followed by a fully connected layer and a Sigmoid activation to get a probability score $p_i = sigmoid(FC_1(f_i))$.
We observe that slices in different slices contributes differently to the identification of GA, which motivates us to implement a location-aware view pooling, illustrated in the right part of Fig.~\ref{fig:multiview_cnn}.
Each slice is assigned to a position id, ranging from $1$ to $100$. The model first uses an embedding layer to embed the position id to a six dimension position feature vector $e_i$. Then, we combine the feature vector $f_i$ extracted from the slice image with the corresponding $e_i$ together. The combined feature vector is fed into a fully connected layer to get the logit score $a_i$. 
 \begin{equation}
     a_i = FC_2( [f_i, e_i])
 \end{equation}
To reduce computational burden during training, we randomly sample $10$ out of the $100$ slices (with an abuse of notation, denoted by $a_1, a_2, \dots, a_{10}$) and send them into a Softmax function to get the attention weights for the $10$ sampled slices, using the following equation
%  \begin{equation}
%      w_i = \frac{exp((a_i + \delta)/\tau)}{\sum_{k=1}^{10} exp((a_k + \delta)/\tau)}
%  \end{equation}
 \begin{equation}
     w_i = \frac{exp((ReLU( a_i ) + \delta)/\tau)}{\sum_{k=1}^{10} exp((ReLU( a_i ) + \delta)/\tau)}
 \end{equation}
Here $\delta$ is a trainable bias term parameter, initialized to a high value to stabilize the training, and gradually attenuated to a small number during training. $\tau$ is the temperature parameter, which is set to a small value to sharpen the attention weight, which helps us to find out the most important slices for GA diagnosis. 
The get final predicted probability of GA (GA score) for an image $x$ at inference time, we compute the weighted summation of the probabilities $w_i$ of all $100$ slices $GA = \sum_i  w_i p_i$. 
% (The model is going to be submitted to an ophthalmology journal.)
% (show model architecture here.)

\subsection{Experiment settings}
The CNN network is an Inception\_v3, which is pre-trianed on ImageNet. For training all models, we use the Adam optimizer with a learning rate of $5 \times 10^{-5}$ with a learning rate decay of $0.5$ every $10$ epochs for the pre-trained CNN network, and the Adam optimizer with a learning rate of $5 \times 10^{-3}$ with a learning rate decay of $0.2$ every $10$ epochs for the other layers in the model.
We use a batch size of $2$ because of the large size of 3D OCT images and our multi-view CNN model (will be illustrated in \ref{sec:multiview}).
Random horizontal flips, and Gaussian noise are used for data augmentations during training.

\section{LIDC Experiment Details}
\label{appendix:lidc}
% \subsection{Dataset description and experiment settings}
We also test the proposed method on a public medical CT scan dataset  LIDC-IDRI~\cite{armato2011lung}.
We follow the settings in~\cite{kohl2018probabilistic, selvan2020tensor} that crops the original images into $128 \times 128$ patches centered on a lesion for which at least one radiologist has annotated.
In our experiment, we focus on the classification task of predicting the presence of a lesions, which is consistent with the setup of~\cite{selvan2020tensor}. There are four radiologists annotates each patch with both lesion label and lesion mask.
A patch in the dataset is labeled as positive if more than two ({\it i.e.} $\geq 3$) radiologists have annotated presence of a lesion, otherwise negative. The ground-truth mask is the pixel-level union set of the four masks.
We use Inception-v3~\cite{szegedy2016rethinking} as our base model for both standard classification and PPI-enhanced training with various saliency mapping schemes.
To match the receptive field of an Inception-v3 model, we resize the input patches to $299 \times 299$. 
For training all models, we use the Adam optimizer with a learning rate of $10^{-4}$ with a learning rate decay of $0.3$ every $10$ epochs after epoch $50$, and a batch size of $64$.
Random horizontal flips, vertical flips and rotations within $20$ degrees are used for data augmentations during training.

% \section{Additional Saliency Map Comparisons}
% Extra saliency map comparisons on CUB, GA, and LIDC are shown in Figure.~\ref{fig:cub_appendix}, \ref{fig:exp_ga_2}, and \ref{fig:lidc_appendix}.

\section{PPI Improving Model Interpretability}
\begin{figure*}[thpb!]
\centering
\includegraphics[width=0.85\textwidth]{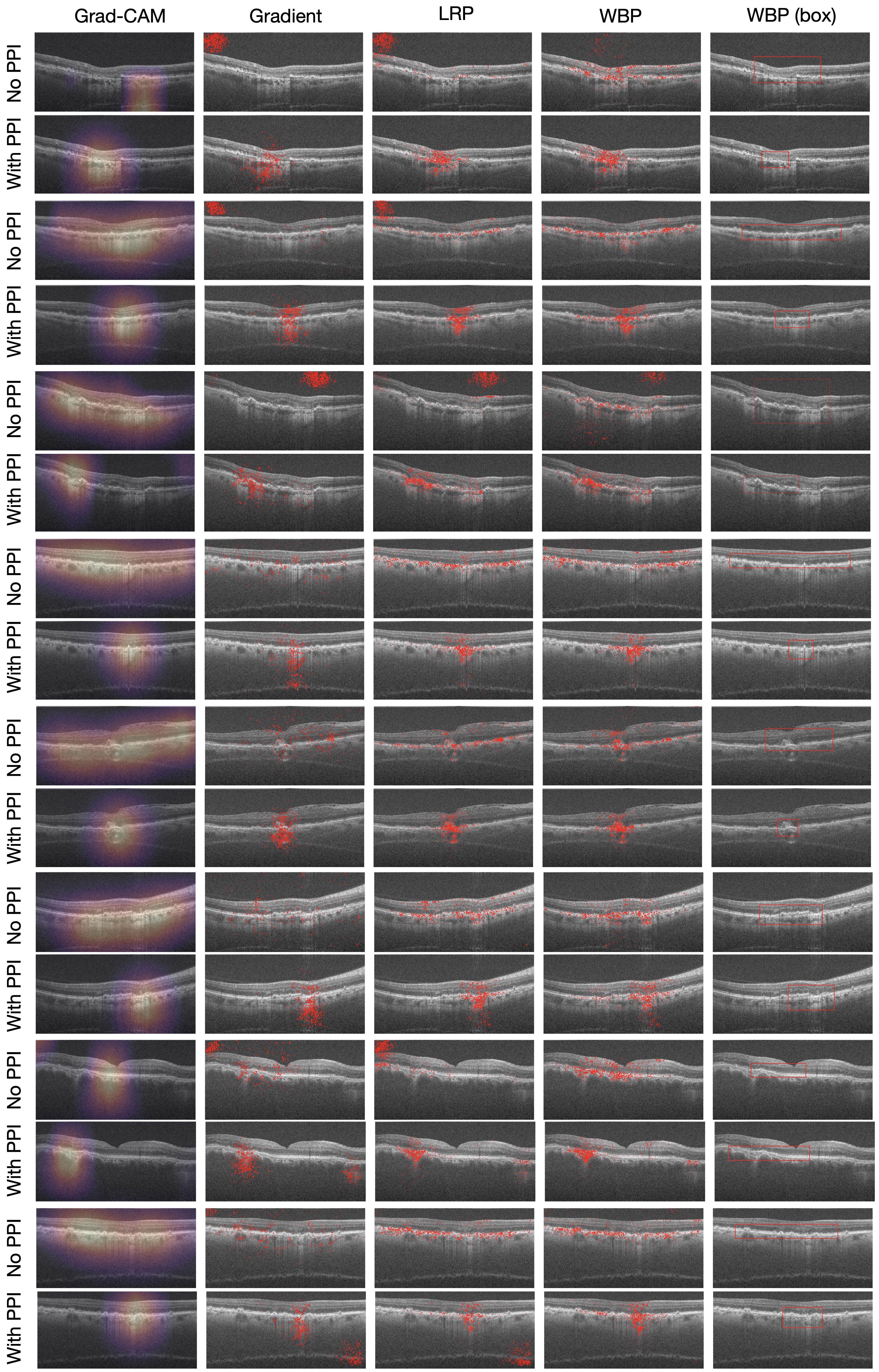} %ga_appendix_more
\caption{Comparison examples of saliency maps on GA dataset based on model trained with and without PPI. } %Each adjacent two rows is a comparison group, the top row shows model trained with standard classification, the row below shows model trained with PPI
\label{fig:exp_ga_2}
\end{figure*}

\begin{figure*}[htpb]
\centering
\includegraphics[width=0.92\textwidth]{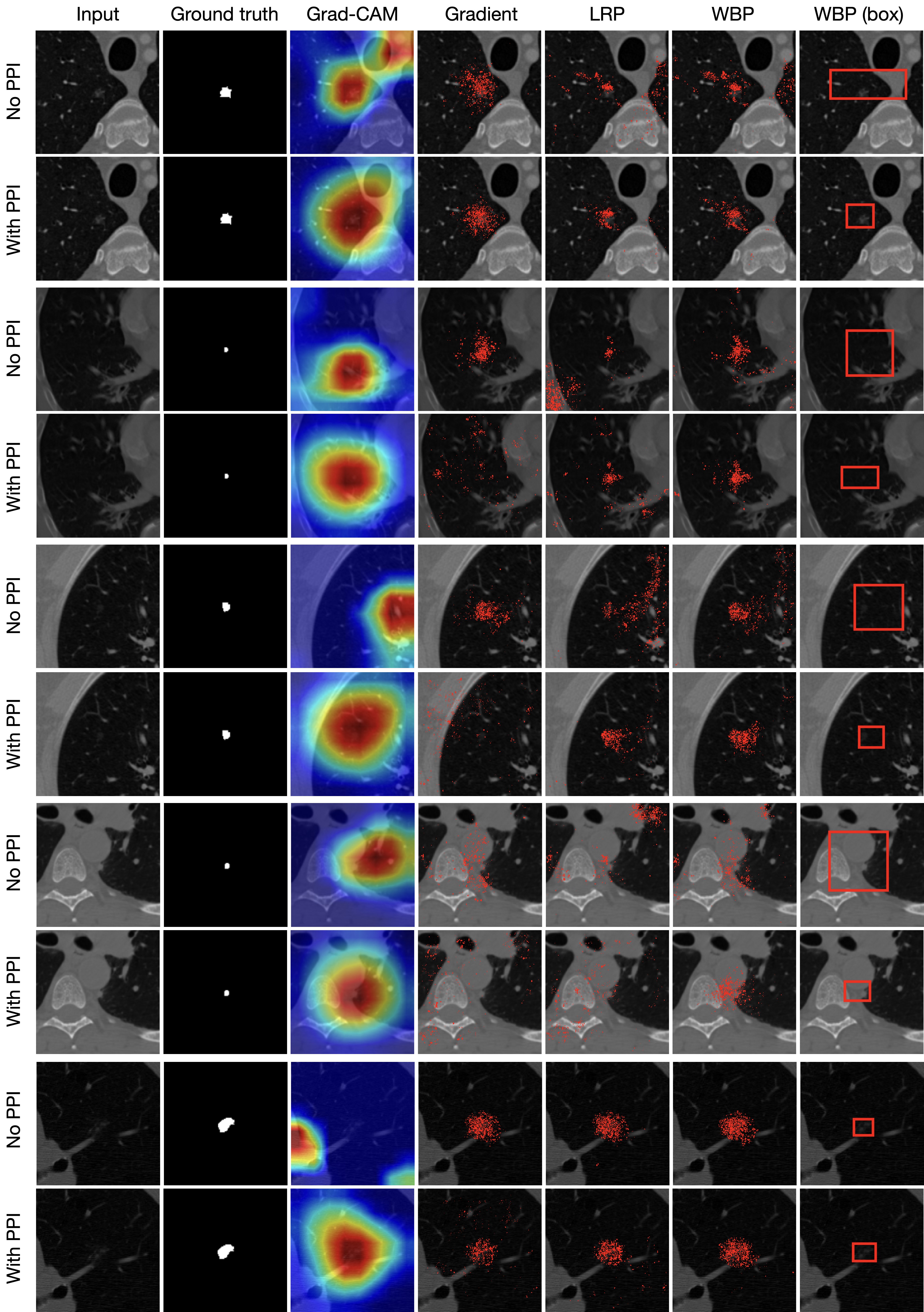}
\caption{Comparison examples of saliency maps on LIDC dataset based on model trained with and without PPI.}
\label{fig:lidc_appendix}
\end{figure*}

Extra comparison examples of saliency maps on GA and LIDC dataset based on model trained without PPI (odd rows) and model trained with PPI (even rows) are shown in Figure.~\ref{fig:exp_ga_2}, and \ref{fig:lidc_appendix}. We find that saliency maps of model trained in PPI learning strategy  are clinically relevant by focusing on retinal layers (GA) or falling on ground-truth segmentations (LIDC).

% \section{Potential application discussion}
% Both quantitative and qualitatively results show that PPI+WBP can not only improve the model performance, but also earn trusts from doctors, which is essential to accelerate clinical deployments of deep learning methods. 

% It is very time consuming and easy to miss small focuses for radiologists to review volume CT scans. The generated fine-grained saliency maps can potentially assistant radiologist to diagnosis scan images by highlighting disease casual related areas. 

% When the classification model outperforms human experts, fine-grained casual saliency maps generated by PPI+WBP can potentially inspire doctors to discover disease related bio-markers, which in turn improves performance of human experts.
% \newpage
% {
%     \bibliographystyle{abbrv}
%     \bibliography{egbib}
% }

\end{document}